%% file: main.tex
\documentclass{article}

\usepackage[table]{xcolor}

\usepackage[final]{corl_2026} 

\usepackage{makecell}
\usepackage{caption}
\usepackage{subcaption}
\usepackage{mwe}
\usepackage{graphicx}
\usepackage{tikz}
\usepackage{array}
\usepackage{multirow}
\usepackage{multicol}
\usepackage{booktabs}
\usepackage{adjustbox}
\usepackage{amsmath}
\usepackage{amssymb}
\usepackage{bm}
\usepackage[cal=boondox, calscaled=.96]{mathalfa}
\usepackage{wrapfig}
\usepackage{float}
\usepackage{enumitem}
\usepackage{pifont}
\usepackage[ruled,linesnumbered]{algorithm2e}
\usepackage{algpseudocode}
\usepackage{listings}
\usepackage[most]{tcolorbox}
\usepackage{etoolbox}

\input{Texs/notations}

\input{Texs/reference_highlights}
\title{\method{}: Open-vocabulary Task-Oriented Dexterous Grasping}

\hypersetup{
  pdftitle={OpenDexGrasp: Open-vocabulary Task-Oriented Dexterous Grasping},
  pdfauthor={Jiyao Zhang, Junhan Wang, Tianyu Wang, Zeyuan Chen, Anthony Bolton, Yitong Peng, and Hao Dong}
}

\author{
  \bfseries Jiyao Zhang$^{1,2*}$ \quad
  Junhan Wang$^{1,2*}$ \quad
  Tianyu Wang$^{1,3*}$\\
  \bfseries Zeyuan Chen$^{1,3}$ \quad
  Anthony Bolton$^{1,2}$ \quad
  Yitong Peng$^{1}$ \quad
  Hao Dong$^{1,2\dagger}$\\[0.5em]
  \normalfont\small
  $^1$ CFCS, School of CS, PKU, China\\
  $^2$ PrimeBot\\
  $^3$ State Key Laboratory of General Artificial Intelligence, BIGAI\\
  $^*$ Equal contribution; $^\dagger$ Corresponding author
}

\begin{document}
\maketitle

\input{Texs/00_overview}
\input{Texs/01_abstract}

\input{Texs/02_introduction}
\input{Texs/03_related_work}
\input{Texs/04_problem_formulation}
\input{Texs/045_opendexverse}
\input{Texs/05_method}
\input{Texs/06_experiments}
\input{Texs/07_conclusion}
\input{Texs/08_limitation}

\acknowledgments{We thank Zhewei Gui and Zimu Han for their insightful discussion. This research was supported by Beijing Natural Science Foundation (26L080330) and National Natural Science Foundation of China (62376006).}

\clearpage

\bibliography{references}  

\clearpage
\appendix
\section*{Appendix}
\input{Texs/09_supplementary}

\end{document}

%% file: Texs/notations.tex
\newcommand{\method}{\textbf{\textsc{OpenDexGrasp}}}
\newcommand{\fulldataset}{\textbf{OpenDexVerse}}
\newcommand{\autodataset}{\textbf{OpenDex-Scale}}
\newcommand{\humandataset}{\textbf{OpenDex-Align}}
\newcommand{\recipe}{\textbf{C2A Recipe}}
\definecolor{RankThirdColor}{RGB}{120,70,160}

\definecolor{CoRLRed}{RGB}{230,100,100}

%% file: Texs/reference_highlights.tex
\newcommand{\AddedReferenceKeys}{}
\forcsvlist{\listadd{\AddedReferenceKeys}}{wu2023graspgf,zhang2025cadgrasp,chen2025clutterdexgrasp,pan2025omnimanip,zhang2026embodied3dbench}

\makeatletter
\let\ODG@original@hyper@\NAT@hyper@
\def\NAT@hyper@#1{%
  \begingroup
  \expandafter\ifinlist\expandafter{\@citeb}{\AddedReferenceKeys}
    {\color{black}\hypersetup{citecolor=black}}{}%
  \ODG@original@hyper@{#1}%
  \endgroup
}

\let\ODG@original@lbibitem\@lbibitem
\def\@lbibitem[#1]#2{%
  \par
  \ifinlist{#2}{\AddedReferenceKeys}
    {\color{black}\hypersetup{urlcolor=black,linkcolor=black}}
    {\color{black}\hypersetup{urlcolor=mydarkblue,linkcolor=mydarkblue}}%
  \ODG@original@lbibitem[#1]{#2}%
}
\makeatother

%% file: Texs/00_overview.tex
\begin{figure}[H]
    \centering
    \vspace{-30pt}
    \includegraphics[width=\textwidth]{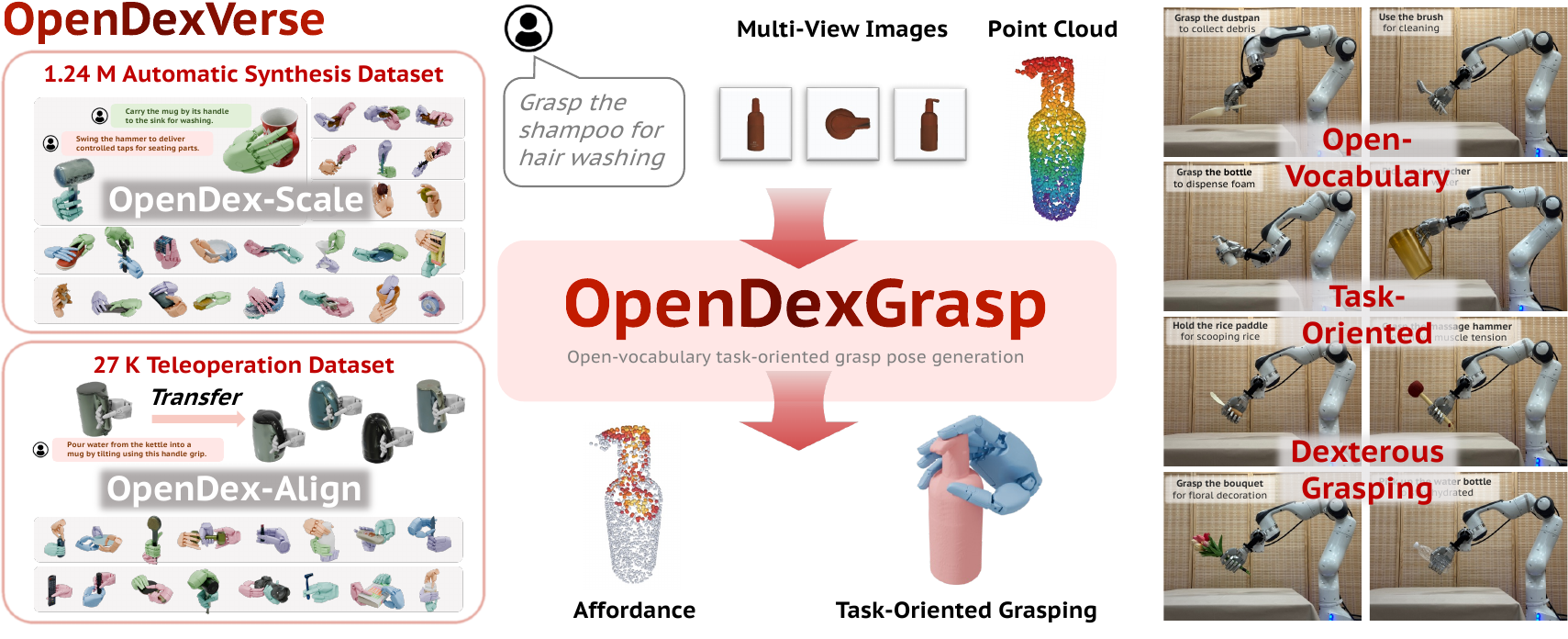}
    \vspace{-15pt}
    \caption{\textbf{Overview of \method{}.} \method{} learns from \fulldataset{} through the \textbf{Coverage-to-Alignment (C2A) Recipe}, coupling large-scale semantic-geometric coverage with high-quality embodied functional alignment. As shown by the real-world demonstrations on the right, \method{} supports \textbf{open-vocabulary} language instructions, grounds \textbf{task-oriented} object functions, and generates \textbf{dexterous grasping} poses for functional manipulation.}
    \vspace{-5pt}
    \label{fig:overview}
\end{figure}

%% file: Texs/01_abstract.tex
\begin{abstract}
Dexterous grasp synthesis has advanced rapidly in generating stable and physically plausible hand poses, but real manipulation requires grasps that preserve the function implied by the task. We study open-vocabulary task-oriented dexterous grasp pose generation, where a robot must infer functional intent from free-form language, ground it in multi-view visual observations and object geometry, and generate an executable high-DoF grasp. We present \method{}, a unified data and generative modeling framework for this setting. \fulldataset{} provides dual-source supervision organized by the Coverage-to-Alignment (C2A) Recipe: \autodataset{} offers large-scale semantic-geometric coverage through automatic grasp synthesis and vision-language annotation, while \humandataset{} supplies high-quality embodied alignment through human teleoperation and category-level transfer. \method{} learns a shared perception-action latent representation that couples open-vocabulary vision-language context with dexterous action generation. Affordance grounding and grasp generation serve as complementary supervision over this latent space, enabling direct generation of task-consistent dexterous grasps without a separate affordance-to-pose inference stage. Extensive simulation and real-robot experiments demonstrate improved functional alignment, physical feasibility, generalization to unseen categories, and real-world execution success. More details and videos are available at \url{https://opendexgrasp.github.io/}.

\end{abstract}

\vspace{-10pt}
\keywords{Robotics, Dexterous Hand, Task-Oriented Grasping}
\vspace{-5pt}

%% file: Texs/02_introduction.tex
\section{Introduction}

Dexterous grasp generation has advanced rapidly through large-scale synthetic data \citep{wang2023dexgraspnet, zhang2024dexgraspnet2, liu2023contactgen, jiang2021grasptta, chen2025bodex}, contact-aware objectives \citep{li2023gendexgrasp, zhao2024graingrasp}, generative modeling \citep{lu2023ugg, weng2024dexdiffuser, xu2024dgt, wu2023graspgf}, and simulation-based filtering. Modern methods can synthesize high-DoF hand poses that are stable, diverse, and physically plausible for a wide range of object geometries. However, real manipulation does not end at holding an object. A robot should grasp a kettle by the handle when pouring, avoid the blade when handing over a knife, and leave the trigger accessible when using a spray bottle \citep{tang2023graspgpt, chen2023taskdex, jian2025gdexgrasp}. These examples expose a basic distinction: task-agnostic grasping asks whether an object can be held, while functional grasping asks whether it is held in a way that preserves the intended action.

We study open-vocabulary task-oriented dexterous grasp pose generation. Given multi-view visual observations, 3D object geometry, and a free-form instruction, the model must generate a high-DoF grasp whose contact, pose, and physical feasibility are all consistent with the task. This setting is challenging because the relevant information lives at different levels of abstraction. The instruction may contain unseen verbs, part names, object categories, or compositional constraints; the functional region is often local and view-dependent; and the hand pose lies in a continuous high-dimensional space subject to contact, penetration, and joint-limit constraints. Prior work has made progress through language-conditioned grasp generation \citep{wei2024dexgys, li2024semgrasp}, open-vocabulary affordance grounding \citep{nguyen2023openad, nguyen2024affpose, deng2021affordancenet, li2024laso}, vision-language part reasoning \citep{liu2023partslip, zhang2026embodied3dbench}, and demonstration-based functional grasping \citep{wang2024dexcap, qin2022dexmv}, but many methods still struggle to preserve the correlation among task semantics, functional contact, and dexterous pose in a single representation. When semantic grounding, affordance prediction, and grasp synthesis are arranged as explicit cascades, errors can propagate across stages and test-time retrieval or optimization often becomes an efficiency bottleneck.

Our key insight is that functional affordance and dexterous grasp pose should be viewed as two supervision views of the same task-conditioned perception-action distribution \citep{tucker1998relations}, rather than as two independent modules in a causal chain. Cognitive studies of affordance and action-oriented perception suggest that object function, task context, and hand action representations are jointly activated and mutually modulated. In robotic grasp generation, this view motivates learning a shared latent space in which language, functional regions, object geometry, and hand pose are optimized together: affordance supervision makes the latent space interpretable at the object surface, while action supervision makes it generative and physically grounded.

This perspective also determines how the data should be built. Optimization-based grasp generation \citep{wang2023dexgraspnet, li2023gendexgrasp, chen2025bodex} can quickly produce large numbers of physically plausible candidates, and vision-language annotation \citep{radford2021clip} can attach open-vocabulary object, part, and task semantics to those candidates. Such data provide broad coverage, but their functional contact may be noisy and their hand configurations may not match natural task-driven choices. Human teleoperation, in contrast, provides reliable functional contact and natural dexterous poses, but it is expensive and difficult to scale. We therefore build \fulldataset{}, a dual-source data ecosystem organized by a \recipe{} (Coverage-to-Alignment). Its large-scale subset, \autodataset{}, is designed for semantic and geometric coverage through automatic grasp synthesis and open-vocabulary annotation. Its high-quality subset, \humandataset{}, is designed for embodied alignment through human teleoperation and category-level transfer. This recipe lets the model first acquire broad open-vocabulary semantic-geometric priors and then align its grasp distribution to reliable functional behavior.

To realize this shared-latent view, we propose \method{}, a generative model for open-vocabulary functional dexterous grasping. As illustrated in Fig.~\ref{fig:method_overview}, multi-view images and the language instruction provide functional context, while the object point cloud supplies the geometry required for contact-rich hand pose generation. Rather than predicting an affordance map and then optimizing a grasp around it, \method{} fuses vision-language, geometry, and action tokens inside a generative action model \citep{black2024pi0, chi2023diffusionpolicy}, allowing task semantics, object geometry, and hand pose to interact before the grasp is produced. A point-level affordance head supervises the same latent representation, making the learned distribution interpretable at the object surface while preserving direct grasp generation at inference time.

Our contributions are threefold:
\begin{itemize}
    \item We formulate open-vocabulary task-oriented dexterous grasp generation and construct \fulldataset{}, a dual-source data ecosystem composed of \autodataset{} for broad semantic-geometric coverage and \humandataset{} for high-quality functional alignment.
    \item We introduce \method{}, a generative model that jointly learns open-vocabulary functional grounding and dexterous grasp pose generation within a shared perception-action latent space.
    \item We propose the \recipe{} for data construction and training, and evaluate it in simulation and on a real dexterous hand, showing improved functional alignment, physical plausibility, inference efficiency, and real-world task execution success.
\end{itemize}

%% file: Texs/03_related_work.tex
\section{Related Work}
\label{sec:related_work}

\textbf{Dexterous grasp generation.}
Geometry-driven dexterous grasping methods synthesize hand poses from object geometry using contact models \citep{li2023gendexgrasp, zhao2024graingrasp, zhang2025cadgrasp}, optimization \citep{wang2023dexgraspnet}, generative models \citep{lu2023ugg, weng2024dexdiffuser, xu2024dgt}, or large-scale synthetic grasp datasets \citep{wang2023dexgraspnet, zhang2024dexgraspnet2, xu2023unidexgrasp}. They have substantially improved physical feasibility and grasp diversity, but their objectives are usually centered on stable holding rather than functional use. Language-conditioned methods \citep{wei2024dexgys, li2024semgrasp} extend this setting by conditioning grasp generation on task descriptions or intent labels. These methods show that language is an effective control signal for dexterous grasping, but many remain limited by relatively closed task spaces or by weak coupling between language semantics and fine-grained hand-object contact patterns.

\textbf{Open-vocabulary affordance and part reasoning.}
Open-vocabulary affordance grounding maps free-form language to object parts, contact regions, or avoid regions \citep{deng2021affordancenet, nguyen2023openad, li2024laso, pan2025omnimanip}. Vision-language and language-model-based approaches can reason about unseen part names and task verbs \citep{liu2023partslip, tang2023graspgpt, kirillov2023sam}, and recent task-oriented grasping systems use these predictions to filter or guide grasp candidates \citep{tang2023graspgpt, chen2023taskdex}. Such pipelines are attractive because their intermediate outputs are interpretable. Their weakness is that the affordance prediction often becomes a hard bottleneck: errors in part grounding propagate to action generation, while test-time retrieval or optimization can become expensive. \method{} keeps the interpretability of point-level affordance supervision \citep{nguyen2024affpose} but uses it to shape a shared action representation rather than as a mandatory inference stage.

\textbf{Functional dexterous grasping and execution-aware evaluation.}
Functional grasping ultimately needs to be judged by whether the generated pose supports downstream manipulation, not only by penetration or lift stability. Demonstration-based \citep{wang2024dexcap, qin2022dexmv} and reinforcement-learning \citep{xu2023unidexgrasp, wan2023unidexgrasp2, chen2025clutterdexgrasp} approaches have made progress on closed-loop functional execution, while contact-map \citep{li2023gendexgrasp} and part-prior methods \citep{jian2025gdexgrasp} improve generalization through structured object-centric representations. \method{} is complementary to these ideas: it combines structured 3D geometry with open-vocabulary visual-language context and evaluates both simulated physical metrics and real-world functional success.

%% file: Texs/04_problem_formulation.tex
\section{Problem Formulation}
\label{sec:problem_formulation}

We formulate open-vocabulary task-oriented dexterous grasping as conditional hand pose generation. Each example is represented by \((u, \mathcal{I}, P, a, m),\) where \(u\) is a free-form functional instruction, \(\mathcal{I}=\{I_v\}_{v=1}^{V}\) denotes multi-view RGB observations of the object, \(P\in\mathbb{R}^{N\times C}\) denotes the object point cloud, \(a\) is a dexterous grasp pose, and \(m\in[0,1]^N\) is an optional point-level functional affordance label. The affordance label is used only as auxiliary supervision; the target task is to directly generate a dexterous grasp pose from language, images, and geometry.

The grasp pose is parameterized as \(a = [p, r_{6D}, q] \in \mathbb{R}^{D}\), where \(p\in\mathbb{R}^{3}\) is the global hand translation, \(r_{6D}\in\mathbb{R}^{6}\) is the continuous 6D wrist rotation representation \citep{zhou2019rotation}, and \(q\in\mathbb{R}^{D-9}\) denotes the hand joint configuration.

The learning objective is to model the conditional distribution over functional dexterous actions, \(p_\theta(a \mid u,\mathcal{I},P),\) so that sampled grasps are both physically feasible and functionally consistent with the instruction. The image-language input provides open-vocabulary task context, while the point cloud provides object geometry for hand-object contact.

Conceptually, we model affordance and action as two observations of a shared latent state \(z\):
\[
    p_\theta(m,a\mid u,\mathcal{I},P)
    =
    \int p_\theta(m\mid z,u,P)\,
         p_\theta(a\mid z,u,P)\,
         p_\theta(z\mid u,\mathcal{I},P)\, dz .
\]
This factorization is not a test-time affordance-to-action cascade. It formalizes the design principle used throughout the paper: affordance grounding and grasp generation should supervise the same perception-action latent state, with affordance providing interpretable surface grounding and action supervision making the latent distribution generative.

%% file: Texs/045_opendexverse.tex
\section{\fulldataset{} and C2A Recipe}
\label{sec:data_recipe}

\subsection{\fulldataset{} Construction}
\label{sec:opendexverse}

\begin{figure}[tbp]
\centering
\includegraphics[width=\textwidth]{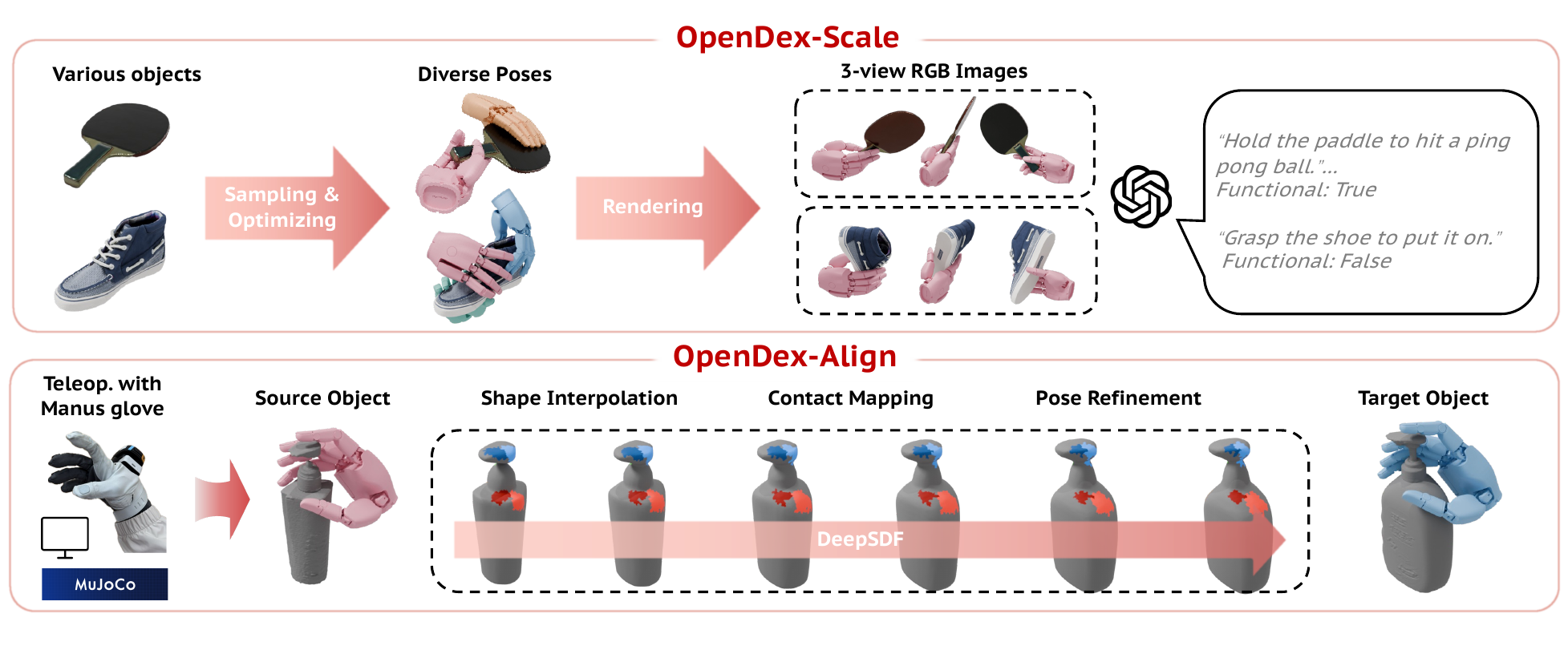}
\caption{\fulldataset{} construction pipeline. Starting from category-aligned real scanned objects, \autodataset{} builds large-scale functional coverage through automatic grasp synthesis and vision-language annotation, while \humandataset{} builds high-quality alignment through teleoperation and dense-correspondence transfer. Together, they instantiate the Coverage-to-Alignment (C2A) Recipe for training \method{}.}
\label{fig:data_statistics}
\end{figure}

\fulldataset{} is a dual-source data ecosystem for learning open-vocabulary functional dexterous grasps. As shown in Fig.~\ref{fig:data_statistics}, it follows the Coverage-to-Alignment (C2A) Recipe. \autodataset{} provides coverage: large-scale automatically synthesized grasps paired with vision-language functional annotations over diverse categories and shapes. \humandataset{} provides alignment: compact but high-quality teleoperated and transferred demonstrations that encode reliable functional contact and natural hand articulation. The two subsets therefore play different roles in the same learning pipeline: first expand the support of language-conditioned grasping, then align that support to embodied functional behavior.

\begin{table}[tbp]
\centering
\setlength{\tabcolsep}{8pt}
\renewcommand{\arraystretch}{1.3}
\input{Tables/data_source}
\caption{\fulldataset{} training data. \autodataset{} provides large-scale semantic-geometric coverage, while \humandataset{} provides embodied functional alignment through human demonstrations and category-level transfer.}
\vspace{-20pt}
\label{tab:data_source}
\end{table}

\paragraph{Object Source.}
We use Omni6DPose~\cite{zhang2024omni6dpose} as the object source because it provides real scanned instances aligned in category-level coordinate spaces. We select categories and instances whose scale, geometry, and functional use are suitable for multi-finger dexterous interaction. This yields an object pool that combines realistic intra-category shape variation with a shared coordinate structure, enabling both automatic grasp synthesis and correspondence-based transfer.

\paragraph{\autodataset{}: Large-Scale Functional Coverage.}
\autodataset{} is designed to cover a broad support of object geometries, grasp modes, and functional language. For each selected object instance, we generate physically plausible dexterous grasp candidates through sampling and optimization following BoDex synthesis pipeline. Each candidate is rendered from three object-centered views and annotated by a vision-language model, which determines whether the grasp preserves a functional use and, when it does, produces the corresponding task description. The resulting data are intentionally broad rather than perfectly curated: they expose the model to diverse semantic-geometric configurations that are difficult to obtain through manual demonstration alone.

\paragraph{\humandataset{}: High-Quality Embodied Alignment.}
\humandataset{} provides the complementary alignment signal. For each selected category-size pair, we choose one representative instance and collect high-quality task-oriented dexterous grasps through human teleoperation. Following \cite{YangCVPR2022OakInk}, we then build dense correspondences within the category-level coordinate space and transfer each demonstration to nearby instances of compatible scale. This produces a smaller but cleaner set of functional grasps, where contact placement, hand posture, and task intent are grounded in embodied demonstrations while still extending beyond the directly teleoperated objects.

\begin{figure}[tbp]
\centering
\includegraphics[width=\textwidth]{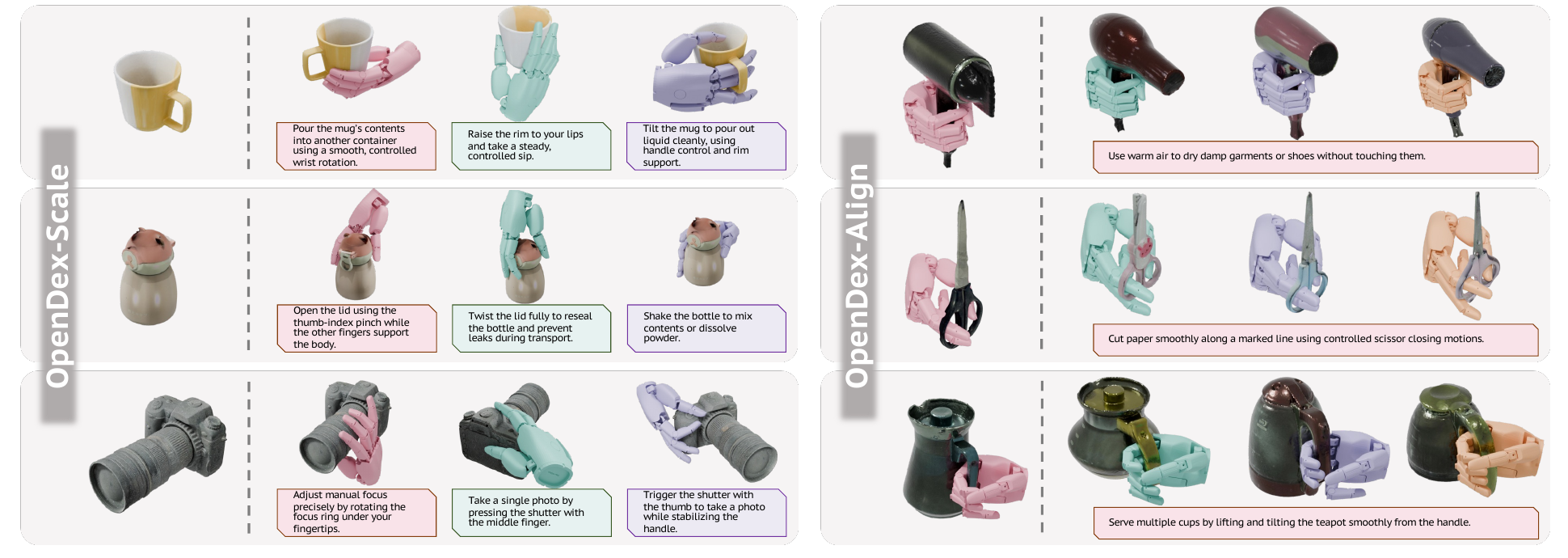}
\caption{\textbf{Examples from \autodataset{} and \humandataset{}.}}
\vspace{-15pt}
\label{fig:dataset_examples}
\end{figure}

\subsection{C2A Details}
\label{sec:c2a}

\method{} is trained with the \recipe{} described above. The recipe separates learning into a coverage stage and an alignment stage while keeping the model architecture and objectives unchanged. In the coverage stage, we pretrain on \autodataset{} using the action-generation and affordance-grounding objectives described in Sec.~\ref{sec:Method}. This stage exposes the model to broad object categories, geometric variations, functional parts, task descriptions, and feasible hand-object configurations, thereby initializing an open-vocabulary perception-action prior.

In the alignment stage, we continue training on \humandataset{}. This changes the data distribution from broad automatically synthesized feasibility toward embodied functional preference. The alignment data improve the model's contact selection and hand articulation because the demonstrations encode human teleoperation and category-level transfer rather than only optimization-derived grasp candidates. Since both stages optimize the same latent model, the alignment stage refines the learned grasp distribution without introducing a separate inference procedure.

During inference, \method{} samples \(a_0\sim\mathcal{N}(0,I)\) and integrates the learned velocity field with a fixed number of Euler steps,
\[
    a_{k+1}
    =
    a_k
    +
    \Delta t\,
    F_\theta(a_k,t_k,z_P,\{H^\ell\}),
    \qquad
    \Delta t=\frac{1}{K}.
\]
The final state \(a_K\) is returned as the dexterous grasp action. Because affordance prediction supervises the shared latent representation during training, inference remains a direct generation process rather than an explicit affordance-then-optimization pipeline.

%% file: Tables/data_source.tex
\begin{adjustbox}{max width=\linewidth}
\begin{tabular}{ccccccc}
\toprule
\textbf{Subset} & \textbf{Source} & \#\textbf{Categories} & \#\textbf{Instances} & \#\textbf{Grasps} & \#\textbf{Functional Poses} & \textbf{Functional Ratio} \\
\midrule
\midrule
\autodataset{} & \makecell{Automatic synthesis\\+ VLM annotation} & 105 & 1.11k & 1.24M & 557.18k & 44.93\% \\
\humandataset{} & \makecell{Teleoperation\\+ Category transfer} & 95 & 2.77k & 27.42k & 18.65k & 68.02\% \\
\bottomrule
\end{tabular}
\end{adjustbox}

%% file: Texs/05_method.tex
\section{Method}
\label{sec:Method}

\vspace{-5pt}
\subsection{Overview}
\vspace{-5pt}

\method{} instantiates the shared-latent formulation in Sec.~\ref{sec:problem_formulation} as a direct generative policy. It extracts open-vocabulary functional context from multi-view images and language, conditions on object geometry from the point cloud, and generates dexterous grasp poses with flow matching \citep{lipman2023flow}. Let \(\{H^\ell\}_{\ell=1}^{L}\) denote selected hidden states from the vision-language encoder and let \(z_P=E_P(P)\) be a point-cloud token. The action model learns a time-dependent velocity field \(F_\theta(x_t,t,z_P,\{H^\ell\}_{\ell=1}^{L})\) that transports Gaussian noise to a task-conditioned dexterous action. Point-level affordance supervision is attached to the same latent representation, grounding the action distribution on functional object regions without turning affordance prediction into a test-time bottleneck.

\vspace{-5pt}
\subsection{Open-Vocabulary Semantic-Geometric Latent}
\vspace{-5pt}

We build the latent representation from complementary semantic and geometric sources. For semantics, all object views and the instruction are formatted as a multimodal sequence and passed through a pretrained vision-language encoder, yielding hidden states \(H^\ell \in \mathbb{R}^{B\times S\times d_{\mathrm{vl}}}\), where \(B\), \(S\), and \(d_{\mathrm{vl}}\) denote the batch size, token length, and hidden dimension. We use hidden states rather than final text responses because they retain correlations among task words, object appearance, and view-dependent part evidence.

For geometry, the object point cloud is represented as \(P=\{(x_i,c_i)\}_{i=1}^{N}\), where \(x_i\in\mathbb{R}^{3}\) is the metric coordinate and \(c_i\in\mathbb{R}^{3}\) denotes RGB channels. A hierarchical point-cloud encoder maps \(P\) to a global geometry token \(z_P \in \mathbb{R}^{B\times 1\times d}\), while intermediate point features are retained for affordance decoding. The resulting semantic tokens provide open-vocabulary task and part cues, and the geometry token conditions grasp generation on object shape for hand-object contact.

\begin{figure}[tbp]
\centering
\includegraphics[width=\textwidth]{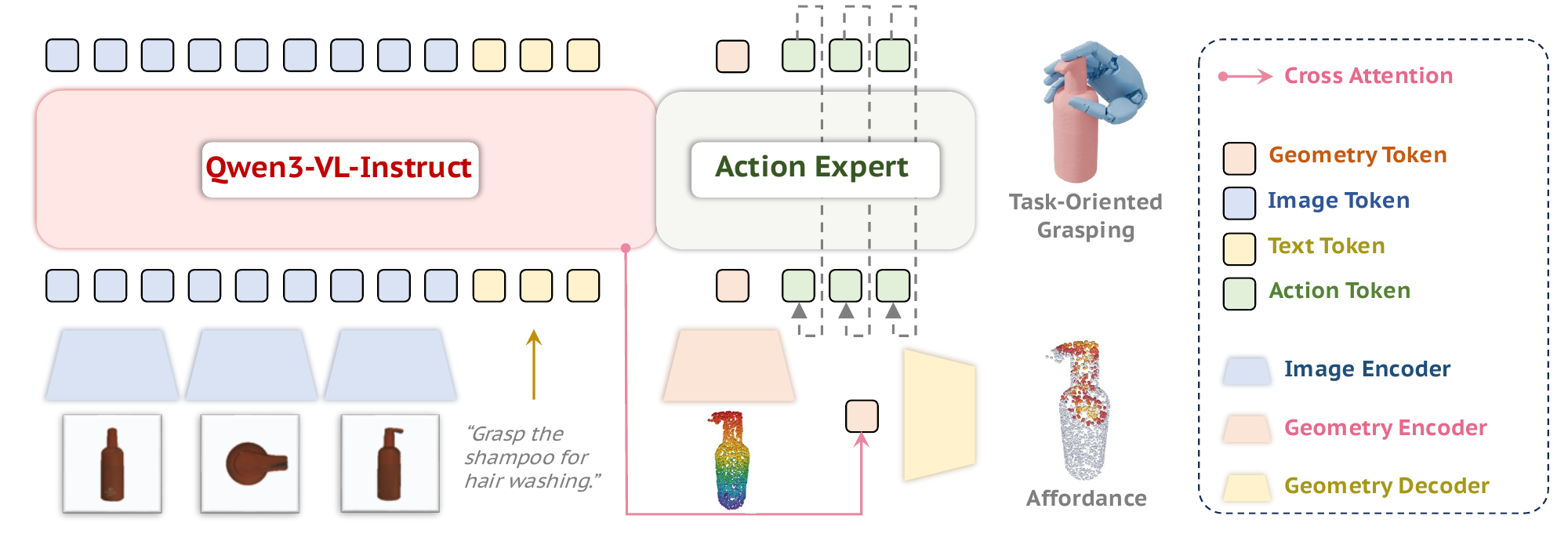}
\caption{\textbf{Architecture of \method{}.} Multi-view images and language provide open-vocabulary semantic tokens, while the point cloud provides geometry tokens. A generative action expert fuses semantic, geometry, and action tokens to directly produce task-oriented dexterous grasps. The affordance head supervises the same latent representation as auxiliary functional grounding, improving interpretability without introducing a separate affordance-to-grasp inference stage.}
\vspace{-15pt}
\label{fig:method_overview}
\end{figure}

\vspace{-5pt}
\subsection{Flow-Matching Dexterous Action Generation}
\vspace{-5pt}

The grasp generator is a transformer-based flow-matching action model. Given a target action \(a\), noise \(\epsilon\sim\mathcal{N}(0,I)\), and time \(t\), we define \(x_t=(1-t)\epsilon+t a\) with target velocity \(v^\star=a-\epsilon\). The noisy action is encoded as action tokens \(z_A\) and concatenated with the geometry token and learned queries, \(S_t=[z_P,z_F,z_A]\). Self-attention fuses geometry, query, and action tokens, while cross-attention conditions them on the vision-language hidden states \(\{H^\ell\}\). The decoder predicts \(\hat{v}_\theta=F_\theta(x_t,t,z_P,\{H^\ell\})\) and is trained with \(\mathcal{L}_{\mathrm{act}}=\mathbb{E}_{a,\epsilon,t}\|F_\theta(x_t,t,z_P,\{H^\ell\})-(a-\epsilon)\|_2^2\). This directly models \(p_\theta(a\mid u,\mathcal{I},P)\), allowing semantic and geometric cues to guide pose generation rather than post-hoc candidate filtering.

\vspace{-5pt}
\subsection{Auxiliary Affordance Grounding}
\vspace{-5pt}

When point-level affordance labels are available, \method{} predicts an auxiliary functional map \(\hat{m}\in[0,1]^{B\times N\times 1}\). The geometry token attends to vision-language features, and the resulting functional condition is propagated to point features to decode surface-level scores. Given affordance labels \(m\), the grounding loss is\(
\mathcal{L}_{\mathrm{aff}}=\mathcal{L}_{\mathrm{focal}}(\hat{m},m)+\lambda_{\mathrm{dice}}\mathcal{L}_{\mathrm{dice}}(\hat{m},m).\)
The full training objective is \(\mathcal{L}=\mathcal{L}_{\mathrm{act}}+\lambda_{\mathrm{aff}}\mathcal{L}_{\mathrm{aff}},\lambda_{\mathrm{aff}}=0.3 .\) 

%% file: Texs/06_experiments.tex
\section{Experiments}
\label{sec:exp}

In this section, we evaluate the proposed method against a strong adapted baseline and conduct comprehensive ablations to validate each design choice. We seek to answer the following questions: \textbf{(1)} Does our method produce higher-quality functional grasps than existing approaches, both on seen and unseen object categories? \textbf{(2)} How does each component of our method contribute to the final performance? \textbf{(3)} Does the learned policy transfer to real-world dexterous manipulation?

\vspace{-5pt}
\subsection{Experimental Setup}
\vspace{-5pt}
\label{subsec:setup}

\textbf{Datasets.} We train on \fulldataset{}, summarized in Table~\ref{tab:data_source}. \autodataset{} provides large-scale automatically generated and semantically annotated grasps covering diverse object categories and geometries. \humandataset{} provides high-quality functional supervision from human teleoperation, and further expands this supervision through category-level transfer to other object instances.

\textbf{Metrics.} We report \textbf{SIV} (Solid Intersection Volume, cm\(^3\)) and \textbf{PD} (Penetration Depth, cm) for hand-object interpenetration, \textbf{SD} (Simulation Displacement, cm) for object center-of-mass displacement after simulation, and \textbf{SR} (Success Rate, \%) for simulated or real task success. \textbf{Style Diversity} is a unitless score computed as the average pairwise distance among repeated stochastic predictions. \textbf{Perceptual Score} (0--10) measures grasp naturalness and task alignment, rated by GPT-5~\citep{openai2025gpt5} and internal lab members.

\textbf{Evaluation protocol.} We separately evaluate \textit{functional} grasps (\textit{func}), where the grasp must support a downstream task (e.g.\ holding the handle of a tool), and \textit{non-functional} grasps (\textit{non-func}), where any stable grasp is acceptable. Both settings are tested on a \textit{seen} split and an \textit{unseen} split, which together measure both fitting and generalization.

\textbf{Baseline.} We compare against \textbf{DexGraspNet 2.0$^\star$}. Since there is no publicly available method under the same open-vocabulary task-oriented dexterous grasping setting as ours, we build this baseline on top of the general-purpose dexterous grasping method DexGraspNet 2.0. The superscript \(\star\) indicates that we augment DexGraspNet 2.0 with CLIP \citep{radford2021clip} features and train and evaluate it on the same datasets and splits as \method{}, enabling a controlled comparison between a strong geometry-driven grasp generator and our task-conditioned functional grasp generator.

\vspace{-5pt}
\subsection{Main Results}
\vspace{-5pt}
\label{subsec:main_results}

\begin{table}[t]
\centering
\input{Tables/main_results}
\caption{Simulation results on seen and unseen splits. \(^{\star}\) denotes our adapted DexGraspNet 2.0 baseline, which adds CLIP features and is trained on the same data as \method{}.}
\vspace{-10pt}
\label{tab:results}
\end{table}

\begin{figure}[tbp]
\vspace{-3pt}
\centering
\includegraphics[width=\textwidth,trim=0 0 0 12,clip]{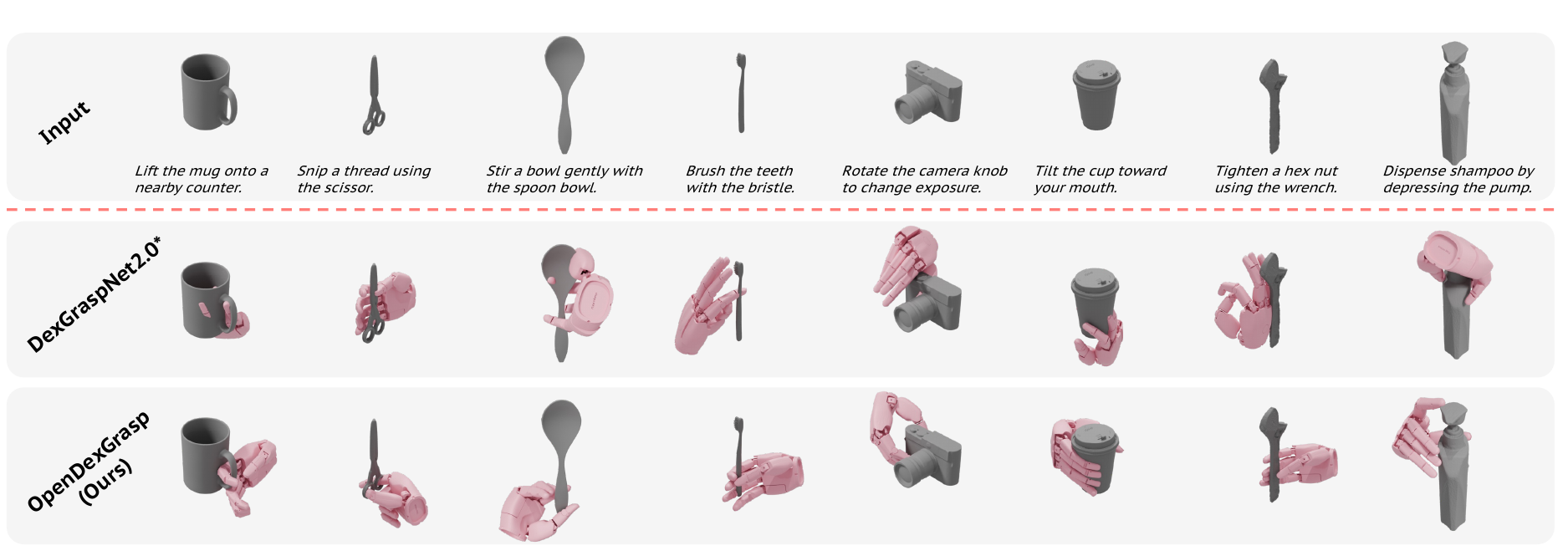}
\caption{Qualitative comparison of predicted grasp poses between \method{} and the adapted DexGraspNet 2.0 baseline. \method{} produces task-consistent dexterous grasps that better align with functional object regions while maintaining plausible hand-object contact.}
\label{fig:pose_qualitative}
\end{figure}

As shown in Table~\ref{tab:results}, \method{} consistently achieves lower SIV, PD, and SD than the baseline, indicating that the produced grasps are physically plausible and remain stable under perturbation. The improvement is especially pronounced on the \textit{unseen} split, where DexGraspNet 2.0$^\star$ suffers a noticeable drop in physical-plausibility metrics, while \method{} maintains a much smaller gap, suggesting stronger generalization to novel object geometries.

In terms of \textit{Style Diversity}, \method{} produces a wider range of grasping styles, demonstrating that the policy does not collapse to a single dominant mode while maintaining stronger physical feasibility. The \textit{Perceptual Score} further evaluates whether this diversity corresponds to natural and task-appropriate dexterous grasping behavior from both automated and human judgments.

\vspace{-5pt}
\subsection{Ablation Study}
\vspace{-5pt}
\label{subsec:ablation}

\begin{table}[t]
\centering
\setlength{\tabcolsep}{8pt}
\renewcommand{\arraystretch}{1.2}
\input{Tables/ablation}
\caption{Ablation results. We report weighted averages over all four simulation splits for the first five metrics and over the functional splits only for Perceptual Score.}
\vspace{-10pt}
\label{tab:ablation}
\end{table}

We conduct ablation studies on the data recipe, functional grounding, and visual-language backbone. Specifically, we remove or weaken four components: \textit{(i)} large-scale \autodataset{} pretraining, \textit{(ii)} high-quality \humandataset{} alignment, \textit{(iii)} affordance grounding, and \textit{(iv)} the pretrained VLM backbone. Results are reported in Table~\ref{tab:ablation}.

\textbf{Effect of large-scale pretraining.} Reducing the scale of \autodataset{} substantially degrades physical robustness and simulation success, with the largest increase in SD among the evaluated ablations. This indicates that large-scale data are critical for learning a broad grasp prior with sufficient geometric and semantic coverage before downstream alignment.

\textbf{Effect of embodied alignment.} Removing the \humandataset{} alignment stage weakens both physical plausibility and perceptual quality. Although the style diversity score slightly increases, the generated variations are less natural and less functionally aligned, suggesting that high-quality teleoperated demonstrations refine the broad pretraining distribution toward reliable contact choices and human-like hand articulation.

\textbf{Effect of affordance grounding.} Without affordance grounding, the model still produces physically plausible grasps, but its simulation success and perceptual quality decrease. This shows that point-level affordance supervision helps the shared representation associate task semantics with functional object regions, rather than merely optimizing for stable hand-object contact.

\textbf{Effect of the pretrained VLM backbone.} In the w/o pretrained VLM setting, we replace Qwen \citep{qwen2025qwen3vl} with CLIP text and image encoders. This variant shows clear degradation across physical, semantic, and success metrics, indicating that a strong pretrained VLM backbone is important for grounding open-vocabulary instructions and object-part cues before dexterous pose generation.

\vspace{-5pt}
\subsection{Real-World Experiments}
\vspace{-5pt}
\label{subsec:real_world}

\begin{table}[t]
\centering
\setlength{\tabcolsep}{4pt}
\renewcommand{\arraystretch}{1.12}
\input{Tables/real_world}
\caption{Real-world results. Success rate and perceptual scores from GPT-5 and human evaluation on five seen and five unseen test objects separately.}
\vspace{-20pt}
\label{tab:human_eval}
\end{table}


To verify transfer beyond simulation, we deploy the learned policy on a real dexterous hand and evaluate it on everyday objects covering both seen and unseen categories. As shown in Table~\ref{tab:human_eval}, \method{} consistently outperforms DexGraspNet 2.0$^\star$ in real-world success rate and perceptual quality under both GPT-5 and internal human evaluation. The gains across familiar and held-out objects suggest that the learned functional grasp policy transfers to physical deployment while preserving task alignment and robust hand-object interaction.

%% file: Tables/main_results.tex
\resizebox{\textwidth}{!}{
\begin{tabular}{c|ccccccc|ccccccc}
\toprule
& \multicolumn{7}{c|}{\textbf{Seen}} & \multicolumn{7}{c}{\textbf{Unseen}} \\
\cmidrule(lr){2-8} \cmidrule(lr){9-15}
& \textbf{SIV}$\downarrow$
& \textbf{PD}$\downarrow$
& \textbf{SD}$\downarrow$
& \makecell{\textbf{SR(\%)} $\uparrow$}
& \makecell{\textbf{Style Div.} $\uparrow$}
& \multicolumn{2}{c|}{\textbf{Perceptual Score} $\uparrow$}
& \textbf{SIV}$\downarrow$
& \textbf{PD}$\downarrow$
& \textbf{SD}$\downarrow$
& \makecell{\textbf{SR(\%)} $\uparrow$}
& \makecell{\textbf{Style Div.} $\uparrow$}
& \multicolumn{2}{c}{\textbf{Perceptual Score} $\uparrow$} \\
\cmidrule(lr){7-8} \cmidrule(lr){14-15}
& & & & & & \textbf{GPT-5} & \textbf{Human} & & & & & & \textbf{GPT-5} & \textbf{Human} \\
\midrule
\midrule
\rowcolor{gray!20}\multicolumn{15}{l}{Functional} \\
\midrule
\textbf{DexGraspNet 2.0$^\star$} & 5.12 & 1.04 & 1.51 & 50.88 & 0.95 & 6.48 & 6.10 & 6.86 & 1.62 & 4.29 & 43.22 & 0.81 & 5.87 & 5.37 \\
\textbf{OpenDexGrasp (Ours)}                    & \textbf{1.28} & \textbf{0.39} & \textbf{1.37} & \textbf{68.07} & \textbf{1.39} & \textbf{7.43} & \textbf{7.85} & \textbf{1.39} & \textbf{0.48} & \textbf{1.80} & \textbf{62.96} & \textbf{1.41} & \textbf{6.98} & \textbf{6.62} \\
\midrule
\rowcolor{gray!20}\multicolumn{15}{l}{Non-Functional} \\
\midrule
\textbf{DexGraspNet 2.0$^\star$} & 4.98 & 0.87 & 2.53 & 63.27 & 1.14 & - & - & 5.16 & 0.99 & 3.87 & 54.49 & 1.03 & - & - \\
\textbf{OpenDexGrasp (Ours)}                    & \textbf{1.33} & \textbf{0.42} & \textbf{0.94} & \textbf{75.46} & \textbf{1.55} & - & - & \textbf{1.30} & \textbf{0.44} & \textbf{2.29} & \textbf{66.73} & \textbf{1.25} & - & - \\
\bottomrule
\end{tabular}
}

%% file: Tables/ablation.tex
\resizebox{\linewidth}{!}{%
\begin{tabular}{lccccccc}
\toprule
\multirow{2}{*}{\textbf{Method}}
& \multirow{2}{*}{\textbf{SIV}$\downarrow$}
& \multirow{2}{*}{\textbf{PD}$\downarrow$}
& \multirow{2}{*}{\textbf{SD}$\downarrow$}
& \multirow{2}{*}{\makecell{\textbf{SR(\%)} $\uparrow$}}
& \multirow{2}{*}{\makecell{\textbf{Style Div.} $\uparrow$}}
& \multicolumn{2}{c}{\textbf{Perceptual Score} $\uparrow$} \\
\cmidrule(lr){7-8}
& & & & & & \textbf{GPT-5} & \textbf{Human} \\
\midrule
\textbf{\method{}} & \textbf{1.31} & 0.41 & \textbf{1.20} & \textbf{71.65} & 1.47 & \textbf{7.37} & \textbf{7.68} \\
\midrule
\quad - w/o \textbf{affordance grounding} & 1.37 & \textbf{0.39} & 1.33 & 67.49 & 1.26 & 6.82 & 6.63\\
\quad - w/o \humandataset{} alignment & 1.58 & 0.53 & 1.44 & 69.38 & \textbf{1.52} & 5.96 & 6.04 \\
\quad - reduced \autodataset{} scale & 2.03 & 0.85 & 9.76 & 60.52 & 1.06 & 6.51 & 6.79 \\
\quad - w/o \textbf{pretrained VLM}  & 5.15 & 0.98 & 2.32 & 57.26 & 1.05 & 6.22 & 6.16 \\
\bottomrule
\end{tabular}
}

%% file: Tables/real_world.tex
\resizebox{\linewidth}{!}{%
\begin{tabular}{ll|ccccc|ccccc|c}
\toprule
\multirow{2}{*}{\textbf{Metric}}
& \multirow{2}{*}{\textbf{Method}}
& \multicolumn{5}{c|}{\textbf{Unseen Category}}
& \multicolumn{5}{c|}{\textbf{Seen Category}}
& \multirow{2}{*}{\textbf{Avg.}} \\
\cmidrule(lr){3-7} \cmidrule(lr){8-12}
& & \makecell{\textbf{Rice}\\ \textbf{Paddle}}
& \textbf{Dustpan}
& \textbf{Bouquet}
& \textbf{Pitcher}
& \makecell{\textbf{Small}\\ \textbf{Bucket}}
& \textbf{Bottle}
& \textbf{Umbrella}
& \textbf{Shampoo}
& \textbf{Hammer}
& \textbf{Brush}
& \\
\midrule
\multirow{2}{*}{\textbf{Success Rate}}
& DexGraspNet 2.0$^\star$
& 40\% & 60\% & 40\% & 50\% & \textbf{50\%}
& 70\% & 80\% & \textbf{60\%} & 80\% & 60\%
& 59.0\% \\
& \textbf{Ours}
& \textbf{60\%} & \textbf{70\%} & \textbf{50\%} & \textbf{80\%} & \textbf{50\%}
& \textbf{100\%} & \textbf{90\%} & \textbf{60\%} & \textbf{90\%} & \textbf{70\%}
& \textbf{72.0\%} \\
\midrule
\multirow{2}{*}{\textbf{GPT-5 Score}}
& DexGraspNet 2.0$^\star$
& 6.2 & 7.1 & 5.5 & 6.7 & \textbf{7.4}
& 7.6 & 6.8 & 6.4 & 7.0 & 6.1
& 6.7 \\
& \textbf{Ours}
& \textbf{8.1} & \textbf{7.2} & \textbf{6.3} & \textbf{7.5} & 7.3
& \textbf{8.6} & \textbf{7.2} & \textbf{6.7} & \textbf{8.3} & \textbf{7.4}
& \textbf{7.5} \\
\midrule
\multirow{2}{*}{\textbf{Human Score}}
& DexGraspNet 2.0$^\star$
& 6.5 & 6.9 & 4.8 & 6.1 & \textbf{7.9}
& 8.0 & 5.9 & 4.8 & 7.6 & 6.8
& 6.5 \\
& \textbf{Ours}
& \textbf{7.7} & \textbf{7.9} & \textbf{7.1} & \textbf{8.0} & 7.1
& \textbf{9.3} & \textbf{7.6} & \textbf{6.8} & \textbf{8.1} & \textbf{7.6}
& \textbf{7.7} \\
\bottomrule
\end{tabular}
}

%% file: Texs/07_conclusion.tex
\section{Conclusion}
\label{sec:conclusion}

We presented \method{}, an open-vocabulary task-oriented dexterous grasping framework that jointly grounds functional affordances and generates hand poses from free-form instructions, multi-view observations, and object geometry. Built on \fulldataset{} and the Coverage-to-Alignment recipe, \method{} combines broad semantic-geometric coverage with high-quality embodied alignment. Simulation and real-robot experiments show improved task alignment, physical plausibility, generalization, and execution success.

%% file: Texs/08_limitation.tex
\section{Limitation}
\label{sec:limitation}

\method{} remains limited by the underlying vision-language representation and visual coverage, so part-grounding errors may affect grasp generation. It also predicts open-loop actions with lightweight execution-time selection, without tactile feedback or closed-loop correction. Finally, \humandataset{} is still modest relative to the diversity of household tools, motivating future work on richer execution data, online feedback, and uncertainty-aware control.

%% file: Texs/09_supplementary.tex

\input{Texs/11_Method_Details.tex}

\input{Texs/12_OpenDexVerse_Construction.tex}

\input{Texs/13_Detailed_Experimental_Setup.tex}

\input{Texs/14_Additional_Experiments.tex}

\input{Texs/15_Failure_Case_Analysis.tex}

%% file: Texs/11_Method_Details.tex
\section{Method Details}
\label{sec:supp_method_details}

\subsection{Implementation Details}
\label{sec:supp_implementation}

\method{} follows the \(\pi_0\)-style vision-language-action architecture~\citep{black2024pi0}, with additional point-cloud encoding and auxiliary affordance grounding modules. The vision-language backbone is initialized from Qwen3-VL-4B-Instruct~\citep{qwen2025qwen3vl} and is executed with FlashAttention-2 under bfloat16 mixed precision. Each training sample consists of rendered object-centered RGB views, a free-form language instruction, a colored object point cloud, and a dexterous grasp action. The action vector parameterizes the wrist translation, wrist rotation, and hand joint values.

The point-cloud encoder adopts a PointNet++ architecture~\citep{qi2017pointnet2}. It takes colored object points as input, where Cartesian coordinates are kept in the original metric scale so that object size remains observable to the model. The encoder uses hierarchical set abstraction followed by global abstraction to produce a global point token. This token is layer-normalized before being passed to the action and affordance modules.

The action head is implemented as a flow-matching diffusion transformer (DiT) with adaptive layer normalization and dropout. The action-side sequence is ordered as \([z_P,z_F,z_A]\), where \(z_P\) is the global PointNet++ token, \(z_F\) denotes learned future-query tokens, and \(z_A\) is the noisy action token. The DiT conditions on selected hidden states from Qwen3-VL. Even-indexed DiT blocks cross-attend to the corresponding Qwen hidden states, while odd-indexed blocks perform causal self-attention over the action-side sequence. The action decoder maps the final action hidden state back to the grasp-action space.

The action head is trained with flow matching. Given a target action \(a_1\), we sample Gaussian noise \(a_0\sim\mathcal{N}(0,I)\) and a continuous time variable \(t\). The interpolated action is \(a_t=(1-t)a_0+t a_1\), and the supervision target is the constant velocity \(v^\star=a_1-a_0\). The training objective is
\[
    \mathcal{L}_{\mathrm{act}}
    =
    \mathbb{E}_{a_0,a_1,t}
    \left[
    \left\|F_\theta(a_t,t,z_P,\{H^\ell\})-(a_1-a_0)\right\|_2^2
    \right].
\]
At inference time, the generated action is obtained by deterministic Euler integration.

The affordance branch predicts point-wise functional scores from the same semantic-geometric latent. The global PointNet++ token is used as a query that cross-attends to selected Qwen hidden states. The resulting text-conditioned geometry tokens are aggregated into a multimodal feature, concatenated into PointNet++ feature-propagation stages, and decoded into per-point affordance scores through a point embedding head followed by a sigmoid classifier. The affordance loss combines focal loss and Dice loss,
\[
    \mathcal{L}_{\mathrm{aff}}
    =
    \mathcal{L}_{\mathrm{focal}}(\hat{m},m)
    +
    \lambda_{\mathrm{dice}}\mathcal{L}_{\mathrm{dice}}(\hat{m},m),
\]
The total objective is \(\mathcal{L}=\mathcal{L}_{\mathrm{act}}+\lambda_{\mathrm{aff}}\mathcal{L}_{\mathrm{aff}}\).

Training is performed with Accelerate and DeepSpeed ZeRO-2 using bfloat16 mixed precision. We optimize the model with AdamW, a cosine learning-rate schedule with warmup, gradient clipping, and a two-stage trainability schedule. The hyperparameters used to train our model are provided in Table~\ref{tab:supp_numeric_hyperparameters}.

\begin{table}[tbp]
\centering
\scriptsize
\setlength{\tabcolsep}{5pt}
\renewcommand{\arraystretch}{1.08}
\begin{tabular}{p{0.42\textwidth}p{0.46\textwidth}}
\toprule
\textbf{Hyperparameter} & \textbf{Value} \\
\midrule
Input views & 3 \\
Point-cloud points & 2048 \\
Action dimension & 31 \\
Wrist translation / rotation / joint dimensions & 3 / 6 / 22 \\
PointNet++ stage-1 samples & 512 \\
PointNet++ stage-1 radii & \([0.1,0.2,0.4]\) \\
PointNet++ stage-1 neighborhood sizes & \([32,64,128]\) \\
PointNet++ stage-2 samples & 128 \\
PointNet++ stage-2 radii & \([0.4,0.8]\) \\
PointNet++ stage-2 neighborhood sizes & \([64,128]\) \\
Global point-token dimension & 1024 \\
DiT blocks & 18 \\
DiT hidden size & 1024 \\
DiT attention heads & 16 \\
DiT dropout & 0.1 \\
Future query tokens & 8 \\
Action horizon & 1 \\
Qwen hidden states for DiT conditioning & 18 \\
Flow-matching time distribution & \(\mathrm{Beta}(2.0,1.0)\) \\
Euler integration steps & 32 \\
Affordance fusion dimension & 1024 \\
Point embedding dimension & 512 \\
Focal loss \(\alpha\) / \(\gamma\) & 0.5 / 2.0 \\
Dice loss weight \(\lambda_{\mathrm{dice}}\) & 1.0 \\
Affordance loss weight \(\lambda_{\mathrm{aff}}\) & 0.3 \\
AdamW betas \((\beta_1,\beta_2)\) & \([0.9,0.999]\) \\
AdamW epsilon & \(10^{-8}\) \\
Weight decay & \(10^{-4}\) \\
Base learning rate & \(3\times10^{-5}\) \\
Qwen-VL interface learning rate & \(5\times10^{-7}\) \\
PointNet++ learning rate & \(3\times10^{-5}\) \\
Action-head learning rate & \(3\times10^{-5}\) \\
Affordance-decoder learning rate & \(3\times10^{-5}\) \\
Optimization steps & 2M \\
Warmup steps & 1k \\
Minimum learning rate & \(3\times10^{-7}\) \\
Per-GPU batch size & 16 \\
Gradient accumulation & 1 \\
Gradient clipping norm & 1.0 \\
Initial frozen-stage duration & 1k steps \\
\bottomrule
\end{tabular}
\caption{Numerical hyperparameters used in \method{}.}
\label{tab:supp_numeric_hyperparameters}
\end{table}

%% file: Texs/12_OpenDexVerse_Construction.tex
\section{OpenDexVerse Construction}
\label{sec:supp_opendexverse}

\subsection{Object Asset Selection}
\label{sec:supp_object_selection}

Our object asset selection pipeline consists of three stages: (1) selecting graspable categories from the Omni6DPose dataset, (2) classifying objects by size within each category, and (3) selecting a subset of representative instances for teleoperation. The first two stages define the object pool used for large-scale \autodataset{} grasp synthesis with BODex, while the third stage selects a smaller subset used for \humandataset{} teleoperation.

\textbf{From Omni6DPose to Graspable Categories.}
The Omni6DPose dataset contains 153 object categories with 4,302 individual instances. We filter these categories based on graspability constraints for dexterous manipulation, focusing on objects that can be effectively grasped with a multi-fingered robotic hand. Categories that are too large (e.g., furniture-scale objects), too flat or thin to afford a stable multi-finger grasp, or otherwise unsuitable for dexterous interaction are removed. This filtering process retains 105 object categories, which serve as the object pool for \autodataset{} grasp synthesis using the BODex pipeline.

\textbf{Size-based Classification.}
To account for size variations within each category, we classify objects based on their maximum diagonal length (the longest distance between any two points on the object mesh). For each category, we compute the standard deviation of diagonal lengths across all instances. Based on the distribution, we classify the 105 categories into three groups:
\begin{itemize}[leftmargin=*]
    \item \textbf{3-class categories} (31 categories): Objects with high size variance are divided into small, medium, and large variants (e.g., ball, bowl, box).
    \item \textbf{2-class categories} (18 categories): Objects with moderate size variance are divided into small and large variants (e.g., bottle, mug, shoe).
    \item \textbf{1-class categories} (56 categories): Objects with low size variance are not further subdivided and labeled as ``all'' (e.g., banana, pen, watch).
\end{itemize}

After size-based classification, the 105 categories expand to 185 category-size pairs: 56 single-size categories ($56 \times 1$), 36 from two-size categories ($18 \times 2$), and 93 from three-size categories ($31 \times 3$). These 185 category-size pairs constitute the full set used for \autodataset{} grasp synthesis.

\textbf{Instance Selection for Teleoperation.}
While all 105 categories (185 category-size pairs) are used for automatic \autodataset{} synthesis, human teleoperation requires objects that a human operator can comfortably manipulate with a dexterous glove. We therefore select a subset of 95 categories (159 category-size pairs) from the 105 categories (185 category-size pairs) for \humandataset{} teleoperation. This selection removes extremely large or awkward instances that are impractical for human teleoperation, while retaining sufficient diversity across object types and sizes. For each selected category-size pair, we choose one representative instance from the Omni6DPose dataset for teleoperation data collection, prioritizing objects with clear geometric features, stable mesh quality, and diverse shape characteristics within their size class.

\textbf{OpenDex-Align: Size-based Transfer.}
For the OpenDex-Align setting, we leverage the size classification to transfer grasp poses across instances within the same category-size pair. Specifically, grasp poses collected from the teleoperated instance are transferred to other instances with similar diagonal lengths (within the same size class). This transfer is performed by scaling and aligning the hand pose relative to the object's normalized coordinate frame, allowing us to efficiently expand the grasp dataset while maintaining grasp quality.

\subsection{Relationship Between Columns in Table 1}
\label{sec:supp_table_columns}

Table 1 in the main paper summarizes the scale and composition of \fulldataset{}, covering both the \autodataset{} and \humandataset{} subsets. This section focuses on the \humandataset{} row and explains how each of its columns is derived and how the numbers relate to each other; the \autodataset{} statistics follow directly from the object pool described in Section~\ref{sec:supp_object_selection}.

\textbf{Category Selection and Size Classification.}
As described in Section~\ref{sec:supp_object_selection}, the 159 category-size pairs used for \humandataset{} teleoperation come from the subset of 95 categories selected for human demonstration.

\textbf{From Category-Size Pairs to Teleoperated Grasps.}
For each of the 159 category-size pairs, we select one representative instance and collect 10 grasp poses through human teleoperation. This results in $159 \times 10 = 1{,}590$ teleoperated grasp poses, which form the \textit{direct teleoperation} portion of \humandataset{}.

\textbf{Category-Level Transfer via Dense Correspondences.}
To expand the teleoperated grasps beyond the directly demonstrated instances, we leverage the category-aligned coordinate space provided by Omni6DPose and build dense shape correspondences following the OakInk methodology. The transfer pipeline consists of eight steps:

\begin{enumerate}[leftmargin=*, label=\textbf{Step \arabic*.}, itemsep=2pt]
    \item \textbf{SDF Preprocessing}: For each category-size pair, we preprocess all object meshes in that class to generate signed distance function (SDF) samples for training.

    \item \textbf{DeepSDF Training}: We train a category-specific DeepSDF model that learns a continuous shape space for all instances within the same category-size pair. Each object is represented by a latent code in this learned space.

    \item \textbf{Shape Reconstruction}: We reconstruct meshes from the learned latent codes to verify the quality of the learned shape representation and ensure sufficient geometric fidelity for grasp transfer.

    \item \textbf{Latent Interpolation}: We generate dense correspondences by interpolating between the source object (teleoperated instance) and each target object in the latent space. This produces intermediate shapes that smoothly morph from source to target geometry.

    \item \textbf{Contact Information Extraction}: For each teleoperated grasp on the source object, we compute hand-object contact information, recording which hand links contact which object surface regions with what contact ratios.

    \item \textbf{Contact Information Transfer}: We transfer the contact information from the source object to each target object by tracking the contact regions through the interpolated shape sequence. As the object geometry morphs from source to target, we identify the corresponding surface points on the target object that should maintain contact with the same hand links.

    \item \textbf{Grasp Pose Refinement}: Given the transferred contact information on the target object, we initialize the hand pose from the source grasp and refine it through gradient-based optimization. The optimization adjusts the hand configuration to maintain the specified contacts on the target geometry while satisfying physical constraints (penetration avoidance, joint limits, self-collision).

    \item \textbf{Penetration Reduction via Simulation}: Because the OakInk transfer pipeline computes penetration metrics on the reconstructed SDF meshes (which differ from the original object geometry), the refined grasp poses may still exhibit penetration when evaluated against the original meshes. To address this discrepancy, we load each transferred grasp into a physics simulator (SAPIEN) with the original object mesh and simulate for 300 steps with position-based drive control. The simulator naturally resolves residual hand-object penetrations through contact dynamics, producing physically plausible transferred grasps that preserve the functional contact configuration.
\end{enumerate}

The transfer is applied within each category-size pair: the 10 teleoperated grasps from the representative instance are transferred to all other instances in the same size class. After filtering out transfers that fail physical validation (e.g., excessive penetration or insufficient contact stability), this process expands the 1,590 directly teleoperated grasps into 25,835 successfully transferred grasps.

\textbf{Summary.}
The columns in Table 1 capture this progression: from the 105 graspable categories and 185 category-size pairs, we select 95 categories (159 category-size pairs) suitable for teleoperation. Each pair yields one teleoperated instance with 10 grasps (1,590 total), and these grasps are transferred to additional instances within each category-size pair via the dense correspondence pipeline. The final \humandataset{} contains $1{,}590 + 25{,}835 = 27{,}425$ grasp poses, combining both teleoperated and transferred grasps to provide high-quality functional supervision across a broad range of object instances.

\subsection{Text Description Generation}
\label{sec:supp_text_generation}

To convert each collected grasp pose into natural language supervision, we design an automated annotation pipeline based on a vision-language model (VLM). For every grasp pose, the pipeline produces three types of labels: (1) a binary \textit{functional / non-functional} classification, (2) five grasp-focused \textit{descriptions}, and (3) five task-focused \textit{instructions}. We use GPT as the annotator, and the same procedure is applied independently to every grasp pose across all object instances.

\textbf{Visual and Contact Inputs.}
For each grasp, we feed the annotator six RGB images together with structured contact information:
\begin{itemize}
    \item \textbf{Three grasp views} rendering the same grasp pose from different viewpoints: a \textit{palm view} from an approximate egocentric perspective looking toward the palm side, a \textit{back view} from a slightly offset angle looking toward the dorsal side of the hand, and a \textit{finger view} that emphasizes the fingertips and their contact geometry with the object.
    \item \textbf{Three object-only reference views} (front, side, and top) of the same object instance, used to convey overall geometry and reveal parts that may be occluded in the grasp views.
    \item \textbf{Hand-object contact information}, pre-computed from hand-link keypoints and object-part meshes. For each grasp, we report which object part is contacted by which hand link, along with the percentage of keypoints on that hand link in contact with the part. This grounds the textual labels in the actual contact configuration rather than visual appearance alone.
\end{itemize}
The annotator is instructed to base the grasp type, contact regions, and functional judgment primarily on the three grasp views and the contact data, while using the object-only views to understand object geometry.

\textbf{Functional Classification.}
The annotator first decides whether the grasp is \textit{functional} or \textit{non-functional}. A grasp is labeled functional when the hand-object configuration enables the object to be used for its intended purpose (e.g., pouring from a bottle, cutting with a knife), as opposed to a non-functional (transport) grasp that merely secures the object for picking up, moving, or handing off. This judgment is grounded in finger placement relative to functional parts, wrist orientation, and whether the grasp affords the object's typical usage motion.

\textbf{Descriptions and Instructions.}
Conditioned on the functional label, the annotator generates five \textit{descriptions} and five \textit{instructions}. Descriptions are grasp-focused and answer ``how is which part of the object being held?'', covering both the grasp type and finger configuration and the specific object part being contacted. Instructions are task-focused and answer ``what task can be done with this hold?'', describing the real-world intent the grasp affords (e.g., driving screws, pouring liquid) without paraphrasing the descriptions. For non-functional grasps, instructions instead describe transport or repositioning tasks. To ensure linguistic diversity, the five descriptions and five instructions are each required to be mutually distinct in wording and focus, with every item phrased as a single concise sentence.

\textbf{Output Format and Quality Control.}
The annotator returns a JSON object containing the boolean \texttt{functional} field together with the \texttt{descriptions} and \texttt{instructions} lists. We post-process each response by stripping list markers and surrounding quotes, normalizing whitespace, and validating that exactly five distinct descriptions and five distinct instructions are present. Responses that fail to parse or violate these constraints trigger an automatic retry with exponential backoff, and we run the pipeline concurrently across many workers to scale annotation to the full dataset.

\textbf{Full Annotation Prompt.}
The complete prompt provided to the annotator is as follows:

\begin{tcolorbox}[breakable, colback=gray!5, colframe=gray!40, boxrule=0.5pt, arc=2mm, left=3mm, right=3mm, top=2mm, bottom=2mm]
\small
\textbf{Goal:} Determine whether the grasp is functional, and generate 5 grasp descriptions and 5 corresponding functional instructions for the given right-hand grasp.

\textbf{Input:} You are given six RGB images related to a robotic hand grasping a \{instance\_name\}. The first three images show the grasp itself: Palm view (first-person perspective toward the palm side), Back view (toward the dorsal side), and Finger view (emphasizing fingertips and contact geometry). The next three images are object-only reference views (Front, Side, Top). Use the object-only views to understand overall geometry; base grasp type, contact regions, and functional judgment primarily on the three grasp images and the contact data.

Along with the images, you are provided with pre-computed contact information between the hand and the object, derived from hand link keypoints and object part meshes. The percentage indicates the proportion of keypoints on that hand link in contact with the object part. Format: \texttt{``Grasp on \textless object\textgreater: \textless object\_part\textgreater: contacted by \textless hand\_link\textgreater(\textless percentage\textgreater\%), ...''}

\textbf{Interpretation:} The image shows one fixed grasp pose performed by the right hand. All outputs must be consistent with the visible hand-object contact, wrist orientation, and object geometry.

\textbf{Constraints:} Do not require regrasping or changing contact regions. Do not invent unseen parts or functions. All outputs must remain physically plausible. Descriptions answer ``How is the object held?'' Instructions answer ``What task can be done with this hold?''

\textbf{Diversity requirement:} The 5 descriptions must differ significantly in wording and focus. The 5 instructions must also differ in functional intent. Avoid repeated sentence structures or near-duplicate meanings.

\textbf{Functional Grasp Classification:} Determine whether this grasp is functional or non-functional (transport). A grasp is functional if the hand-object configuration enables using the object for its intended purpose (e.g., pouring from a bottle, cutting with a knife), not just holding or transporting it. A grasp is non-functional if it merely secures the object for picking up, moving, or handing off. Base your judgment on finger placement relative to functional parts, wrist orientation, and whether the grasp affords the object's typical usage motion. Return true for functional, false for non-functional. If non-functional, instructions should describe transport or repositioning tasks rather than object-use tasks.

\textbf{Descriptions (grasp-focused):} Each description MUST cover both: (1) how the hand grasps—grasp type, finger configuration, force pattern; AND (2) which part of the object is contacted. Think of it as answering ``How is which part of the object being held?'' Do NOT describe task goals or intended use. Bad example: ``The hand firmly holds the object.'' (missing part + configuration detail). Good example: ``A power grasp wraps all four fingers around the cylindrical handle, with the thumb opposing across the shaft.''

\textbf{Instructions (intent-focused):} Describe the real-world task this grasp is suited for—what the person is trying to accomplish. Focus on task-level intent (e.g., driving a screw, pouring liquid, opening a door), not low-level motion. Must be achievable with the exact shown grasp without regrasping. Do NOT repeat or paraphrase content from descriptions. Bad example: ``Grip the handle tightly to hold the object.'' (describes grasp, not task). Good example: ``Use this grip to drive screws into a surface with controlled rotational force.''

\textbf{Length and style constraint (IMPORTANT):} Each item must be ONE short sentence (10–20 words preferred). Be concise and information-dense. Avoid explanations, subordinate clauses, or filler words. Use compact phrasing (like captions, not full explanations). Use varied sentence structures (no fixed template). Each sentence should emphasize one main aspect. English only.

\textbf{Output format:} Return JSON only: \texttt{\{"functional": true, "descriptions": ["...", "...", "...", "...", "..."], "instructions": ["...", "...", "...", "...", "..."]\}}
\end{tcolorbox}

\subsection{Data Visualization}
\label{sec:supp_data_visualization}

\begin{figure}[tbp]
\centering
\includegraphics[width=\textwidth,height=0.9\textheight,keepaspectratio]{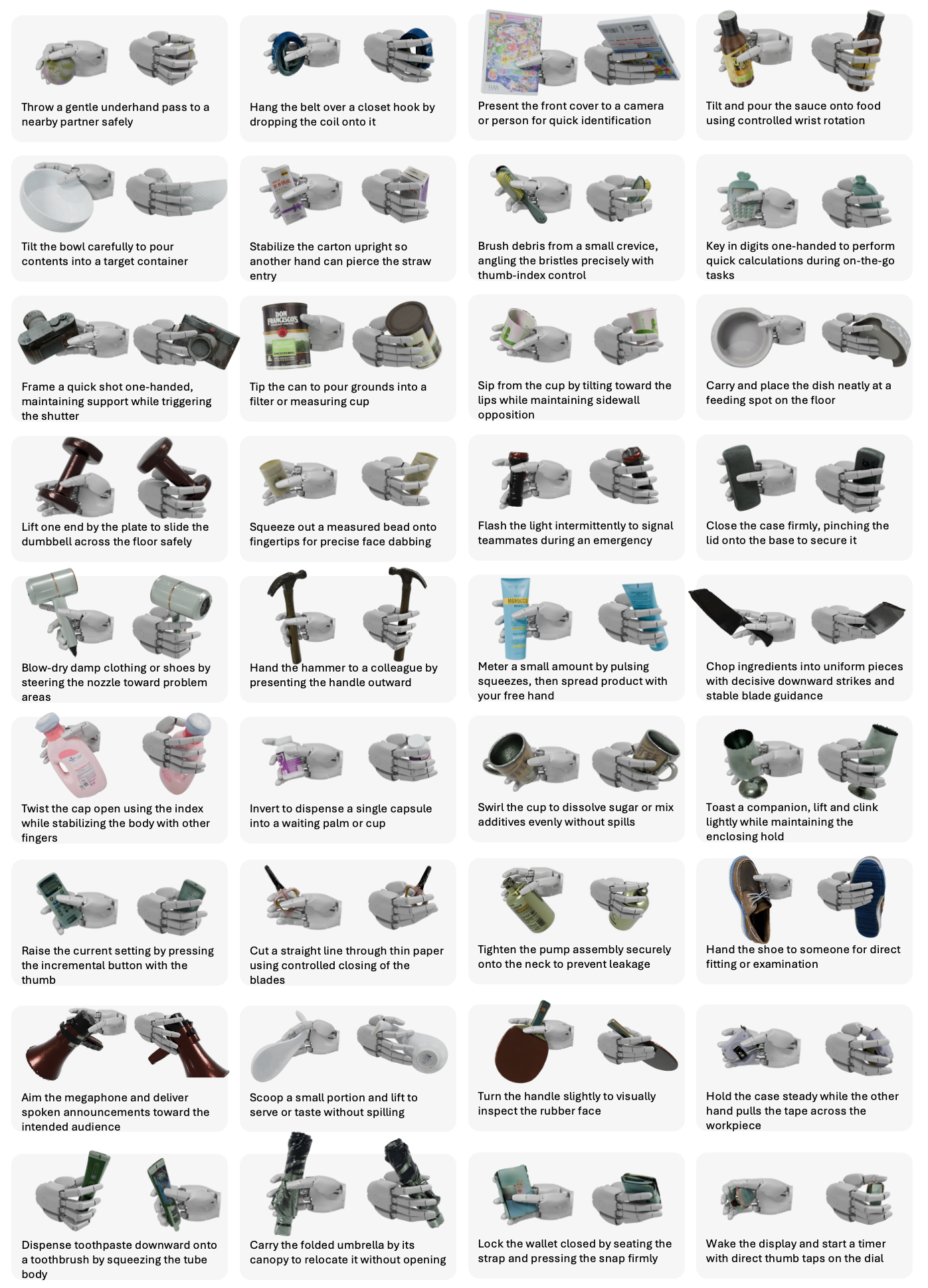}
\caption{\textbf{Visualization of \autodataset{} grasps.} Representative dexterous grasp poses automatically synthesized with the BODex pipeline across diverse object categories and sizes. The broad coverage of geometries, grasp modes, and contact configurations provides the large-scale semantic-geometric support of \fulldataset{}.}
\label{fig:supp_bodex_data}
\end{figure}

\begin{figure}[tbp]
\centering
\includegraphics[width=\textwidth,height=0.9\textheight,keepaspectratio]{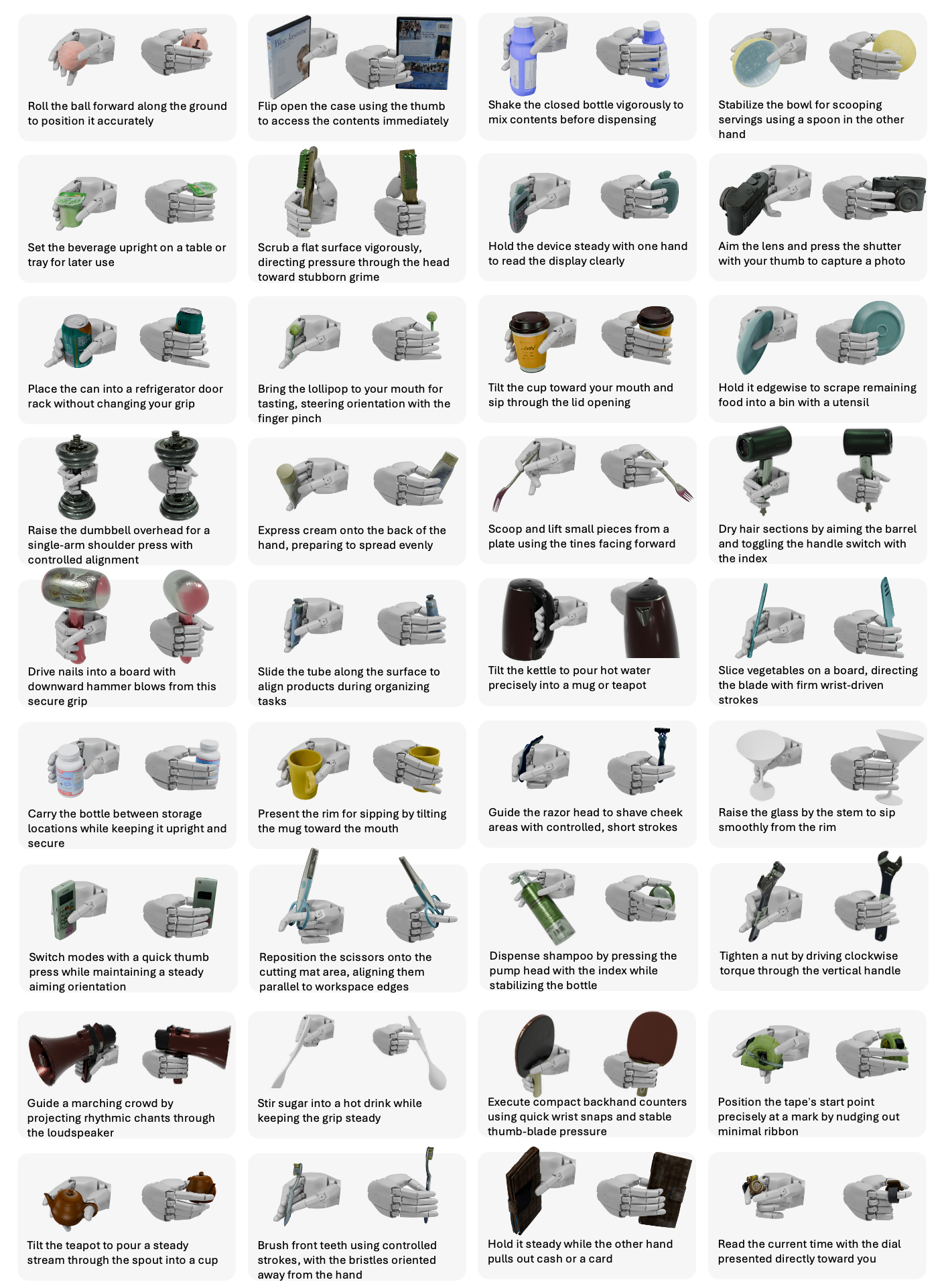}
\caption{\textbf{Visualization of \humandataset{} grasps.} Representative task-oriented dexterous grasps collected through human teleoperation and expanded via dense-correspondence transfer. The natural hand articulation and functional contact placement provide the high-quality embodied alignment signal of \fulldataset{}.}
\label{fig:supp_teleop_data}
\end{figure}

To provide an intuitive sense of the grasp quality and diversity in \fulldataset{}, we visualize representative grasp poses from both data sources. Figure~\ref{fig:supp_bodex_data} shows examples from \autodataset{}, the large-scale coverage subset synthesized with the BODex pipeline, and Figure~\ref{fig:supp_teleop_data} shows examples from \humandataset{}, the high-quality alignment subset collected through human teleoperation and expanded via dense-correspondence transfer.

\textbf{\autodataset{} Visualization.}
Figure~\ref{fig:supp_bodex_data} presents a gallery of automatically synthesized grasps across a broad range of object categories and sizes. Each grasp is generated through sampling and optimization following the BODex synthesis pipeline, yielding physically plausible hand-object configurations without human supervision. The examples illustrate the breadth of semantic-geometric coverage in \autodataset{}: the same object category is grasped under multiple poses and contact modes, and the dataset spans objects of widely varying shape, scale, and topology. This diversity is precisely what enables \autodataset{} to expand the support of language-conditioned grasping beyond what manual demonstration alone could provide.

\textbf{\humandataset{} Visualization.}
Figure~\ref{fig:supp_teleop_data} presents representative grasps from \humandataset{}, collected through human teleoperation with a dexterous glove and then transferred to nearby instances within each category-size pair via dense shape correspondences. Compared with the automatically synthesized grasps, these demonstrations exhibit more natural hand articulation and functionally meaningful contact placement, reflecting how a human operator would actually use each object. The examples highlight the alignment role of \humandataset{}: although smaller in scale, it grounds grasp posture, contact regions, and task intent in embodied demonstrations, providing the high-quality functional supervision that aligns the model's broad coverage to reliable real-world behavior.

%% file: Texs/13_Detailed_Experimental_Setup.tex
\section{Detailed Experimental Setup}
\label{sec:supp_exp_setup}

\subsection{Train/Test Split}
\label{sec:supp_train_test_split}

Our test set is divided into functional grasping and non-functional grasping categories, with each further split into seen and unseen object categories. Functional object categories are those we identify as requiring task-oriented grasping to support specific downstream actions (e.g., holding a knife by the handle for cutting), while non-functional object categories are those that only require general pick-and-place manipulation. For each object category, we selected 3 instances for evaluation. Table~\ref{tab:test_split} summarizes the number of categories and instances in each test split, along with the specific object categories.

\begin{table}[h]
\centering
\small
\begin{tabular}{lcc>{\raggedright\arraybackslash}p{0.48\textwidth}}
\hline
\textbf{Test Split} & \textbf{\# Cat.} & \textbf{\# Inst.} & \textbf{Object Categories} \\
\hline
Functional - Seen & 25 & 75 & bottle, dish, bowl, facial cream, brush, camera, hair dryer, can, hammer, hand cream, remote control, knife, rubik cube, lipstick, scissor, mouse, shampoo, mug, shoe, table tennis bat, tape measure, umbrella, teapot, toothbrush, toothpaste \\
\hline
Functional - Unseen & 5 & 15 & spoon, cup, dumbbell, spanner, whistle \\
\hline
Non-functional - Seen & 46 & 138 & banana, bread, ball, cake, belt, calculator, book, candy, box, carrot, corn, flower pot, glasses case, donut, hamburger, egg, handbag, egg tart, hot dog, keyboard, onion, lemon, orange, mango, peach, mangosteen, pear, medicine bottle, pie, pitaya, soap, pizza, sweet potato, pomegranate, timer, power strip, tissue, sausage, tomato, toy boat, toy train, toy bus, toy truck, toy car, toy motorcycle, toy plane \\
\hline
Non-functional - Unseen & 5 & 15 & toy animals, chili, chicken leg, wallet, frisbee \\
\hline
\textbf{Total} & \textbf{81} & \textbf{243} & \\
\hline
\end{tabular}
\caption{Test set composition across functional and non-functional grasping tasks with seen and unseen object categories.}
\label{tab:test_split}
\end{table}

\subsection{Metric Calculation}
\label{sec:supp_metrics}

We evaluate generated grasps from three complementary aspects: physical plausibility, style diversity, and task-semantic alignment. For physical plausibility, we follow the metrics commonly used in language-conditioned and generalizable dexterous grasping evaluation~\citep{li2024semgrasp,jian2025gdexgrasp}.

\paragraph{\textbf{Penetration Depth}.}
Penetration Depth (PD, cm) measures the maximum distance by which hand vertices penetrate the object surface. Lower PD indicates less severe local interpenetration.

\paragraph{\textbf{Solid Intersection Volume}.}
Solid Intersection Volume (SIV, cm\(^3\)) measures volumetric hand-object intersection. To compute SIV precisely, we voxelize both the hand and object meshes with a \(1\,\mathrm{mm}^3\) voxel unit and sum the volume of voxels lying inside the overlapping solid region.

\paragraph{\textbf{Success Rate}.}
Success Rate (SR, \%) evaluates grasp stability in MuJoCo simulation. For each predicted grasp, the hand is placed at the predicted pose and held fixed. We set the object mass to \(0.1\,\mathrm{kg}\), the friction coefficient to 1.0, enable gravity, and run the simulation rollout for \(2\,\mathrm{s}\). A predicted grasp is counted as successful when the resulting object displacement is below \(0.2\,\mathrm{m}\). For each test sample, \method{} is evaluated with 10 stochastic inference trials. Let \(N\) be the number of test samples and \(s_{i,j}\in\{0,1\}\) denote whether the \(j\)-th trial for sample \(i\) succeeds. The simulated success rate is
\[
    \mathrm{SR}
    =
    \frac{1}{10N}
    \sum_{i=1}^{N}
    \sum_{j=1}^{10}
    s_{i,j}
    \times 100\%.
\]

\paragraph{\textbf{Simulation Displacement}.}
Simulation Displacement (SD, cm) is computed only for grasps that are judged successful under the Success Rate protocol. For these successful grasps, SD uses the same MuJoCo rollout to measure the displacement of the object center of mass after simulation. Lower SD indicates a more stable grasp.

\paragraph{\textbf{Style Diversity}.}
Style Diversity measures the variability among repeated stochastic predictions for the same test sample. For each test sample \(i\), \method{} performs 10 inference trials and produces predicted poses \(\{(\mathbf{t}_{i,k}, \mathbf{R}_{i,k}, \mathbf{q}_{i,k})\}_{k=1}^{10}\), where \(\mathbf{t}_{i,k}\) is the wrist translation, \(\mathbf{R}_{i,k}\) is the wrist rotation, and \(\mathbf{q}_{i,k}\in\mathbb{R}^{22}\) is the hand joint-angle vector. For each pair of predictions, we compute the wrist translation difference in meters, the wrist rotation difference in axis-angle magnitude in radians, and the 22-DoF hand joint difference in radians, and weight the three terms equally:
\[
    d_{i,a,b}
    =
    \frac{1}{3}\left(
    \|\mathbf{t}_{i,a}-\mathbf{t}_{i,b}\|_2
    + \|\operatorname{Log}(\mathbf{R}_{i,a}^{\top}\mathbf{R}_{i,b})\|_2
    + \|\mathbf{q}_{i,a}-\mathbf{q}_{i,b}\|_2
    \right).
\]
\[
    \mathrm{StyleDiversity}_i
    =
    \frac{1}{\binom{10}{2}}
    \sum_{1\leq a<b\leq 10}
    d_{i,a,b}.
\]
The reported Style Diversity is averaged over all test samples.

For semantic alignment, we use both GPT-5-assisted evaluation~\citep{openai2025gpt5} and human rating. For each generated grasp, we render the hand-object interaction from multiple views together with the input instruction. GPT-5 is asked to score whether the grasp preserves the functional region implied by the instruction and whether the hand pose is natural for the specified task. Scores range from 0 to 10, where 0 means the grasp is entirely inconsistent with the instruction and 10 means the grasp is both physically plausible and functionally aligned. For human evaluation, we randomly sample 30 groups of results and ask 20 participants to rate grasp quality and task alignment on the same 0--10 Likert scale. The reported human perceptual score is the mean rating across participants and sampled groups.

\subsection{Baseline Details}
\label{sec:supp_baseline}

There is no publicly available method that directly addresses our open-vocabulary task-oriented dexterous grasping setting. We therefore construct a strong adapted baseline, denoted DexGraspNet 2.0\(^{\star}\), by augmenting DexGraspNet 2.0~\citep{zhang2024dexgraspnet2} with language conditioning. The baseline preserves the original two-stage DexGraspNet 2.0 pipeline, including seed-point proposal and grasp-pose generation, and is trained and evaluated on the same data splits as \method{}.

Given a colored object point cloud, the ResUNet14 backbone extracts point-wise local geometric features. To incorporate language instruction, we encode each instruction with a CLIP text encoder~\citep{radford2021clip} and obtain a sentence-level language embedding. The language embedding is concatenated with each point-wise feature, producing language-aware local features for all object points. These fused features are used to predict language-conditioned graspness scores, from which high-scoring object points are selected as seed points. The selected seed features then condition the downstream grasp generator. Specifically, the diffusion-based wrist-pose generator predicts the wrist-pose distribution conditioned on the fused language-geometry feature, while the joint-angle prediction MLP estimates the hand joint configuration from the generated wrist pose and the same fused feature. This design enables the baseline to select task-relevant grasp regions and generate language-aligned dexterous grasps, providing a controlled comparison with our unified open-vocabulary perception-grasping model.

\subsection{Real Robot Setup}
\label{sec:supp_robot_setup}

\begin{figure}[tbp]
\centering
\includegraphics[width=0.82\textwidth]{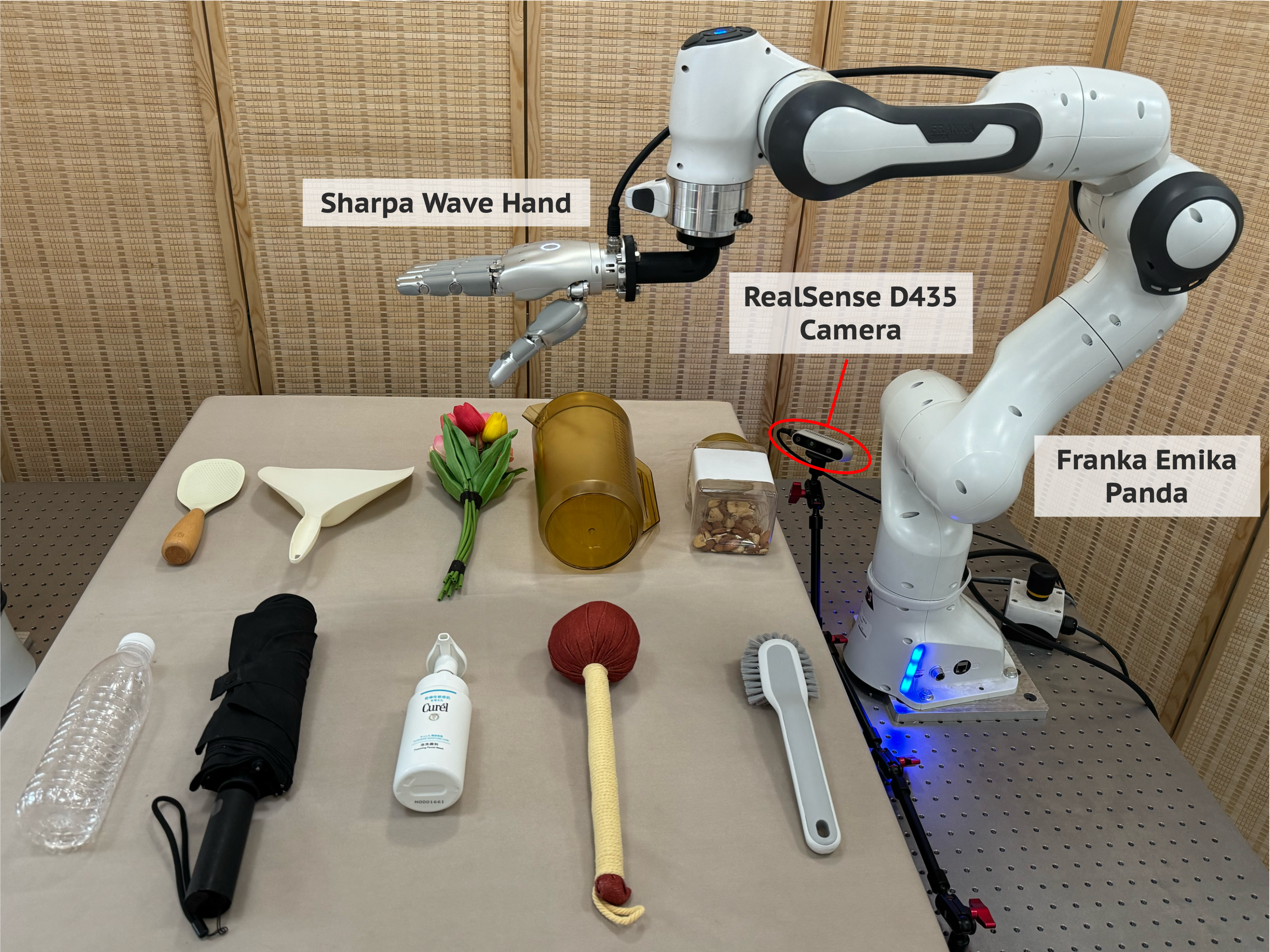}
\caption{Real-world experiment setting. The Sharpa Wave Hand is mounted on a Franka Emika Panda robotic arm, with an Intel RealSense D435 camera used for object pose estimation before grasp execution.}
\label{fig:real_world_experiment_setting}
\end{figure}

As shown in Fig.~\ref{fig:real_world_experiment_setting}, our real-world experiments are conducted with a Sharpa Wave Hand mounted on a Franka Emika Panda robotic arm. An Intel RealSense D435 camera is mounted near the robot base and is used to estimate the 6D pose of the target object before grasp execution. At test time, \method{} receives the reconstructed object point cloud, multi-view observations, and a language instruction specifying the intended functional use, such as ``use the brush for cleaning.'' The model predicts a dexterous grasp pose, which is then transformed into the robot frame and executed with motion planning.

For each object and its corresponding language instruction, we repeatedly run inference with \method{} to sample candidate grasp poses. After each inference run, we manually discard the predicted grasping pose that would collide with or be occluded by the table, since such poses cannot be safely reached by the physical robot. We continue this process until 10 valid predicted poses are collected for each object. To account for different friction properties across real objects, we apply a fixed squeezing refinement after reaching the predicted grasp: each hand joint is increased by \(0.05\,\mathrm{rad}\) along the closing direction. We then deploy all 10 predictions in the real-world setup, executing each pose once. We report real-world SR as the percentage of trials in which the robot successfully grasps and lifts the target object. Due to hardware safety constraints, we do not report real-world trials on objects whose shape or required motion would create a high risk of collision or unstable manipulation.

%% file: Texs/14_Additional_Experiments.tex
\section{Additional Experiments}
\label{sec:supp_additional_exp}

\subsection{Ablation Study: w/o Pose Generation Module}
\label{sec:supp_ablation}

A central design choice of our method is to \emph{directly generate} dexterous grasp poses through a learned generative model, rather than first predicting an affordance map and then \emph{optimizing} a grasp pose to satisfy it. To validate this choice, we construct an optimization-based variant that removes the pose generation module while keeping the rest of our pipeline unchanged. This variant follows the affordance-then-optimization paradigm popularized by GenDexGrasp~\cite{li2023gendexgrasp}: it consumes the same point-level affordance predicted by our shared latent representation, but instead of decoding a grasp pose, it solves for the hand configuration via gradient-based optimization.

\textbf{Optimization-based Baseline.}
Given the predicted affordance over the object point cloud, we optimize the 28-dimensional hand configuration (3 for wrist translation, 3 for wrist rotation, and 22 for finger joints) using Adam. The hand is initialized along the direction of the highest-affordance point with the palm facing the object center. At each step, we minimize a weighted sum of five energy terms:
\begin{itemize}[leftmargin=*, itemsep=2pt]
    \item \textbf{Penetration energy} that pushes hand points out of the object surface using point-to-surface signed projections along object normals.
    \item \textbf{Affordance attraction energy} that pulls finger anchors toward high-affordance regions using an alignment-aware distance, where the alignment term down-weights contacts that approach the surface from a grazing angle.
    \item \textbf{Coverage energy} that encourages every high-affordance cluster to be reached by at least one finger anchor, preventing the hand from collapsing onto a single contact region.
    \item \textbf{Joint-limit energy} that penalizes violations of the hand's joint limits.
    \item \textbf{Self-penetration energy} that penalizes inter-link collisions using a sphere-based collision model.
\end{itemize}
We run optimization for up to 300 steps per grasp with a plateau-based learning-rate schedule. This baseline shares the identical affordance input, object geometry, and hand kinematic model as our method, isolating the effect of replacing direct pose generation with test-time optimization.

\textbf{Results.}
Table~\ref{tab:ablation_pose_gen} compares our method against this optimization-based variant using the same metrics as our main experiments: Solid Intersection Volume (SIV), Penetration Depth (PD), Simulation Displacement (SD), Success Rate (SR), Style Diversity, and Perceptual Score rated by GPT-5 and human evaluators. We additionally report the average inference time per grasp. Our generative pose module not only produces grasps with better physical plausibility, higher functional success, and superior perceptual quality, but also runs substantially faster: it generates a grasp pose in a fraction of a second using 32 steps of Euler integration, whereas the optimization baseline requires hundreds of iterative steps per grasp. This confirms that learning to directly generate grasp poses, rather than optimizing against a predicted affordance at test time, yields both better quality and far lower inference latency.

\begin{table}[h]
\centering
\small
\setlength{\tabcolsep}{5pt}
\renewcommand{\arraystretch}{1.2}
\resizebox{\textwidth}{!}{%
\begin{tabular}{lcccccccc}
\toprule
\multirow{2}{*}{\textbf{Method}}
& \multirow{2}{*}{\textbf{SIV}$\downarrow$}
& \multirow{2}{*}{\textbf{PD}$\downarrow$}
& \multirow{2}{*}{\textbf{SD}$\downarrow$}
& \multirow{2}{*}{\makecell{\textbf{SR(\%)} $\uparrow$}}
& \multirow{2}{*}{\makecell{\textbf{Style Div.} $\uparrow$}}
& \multicolumn{2}{c}{\textbf{Perceptual Score} $\uparrow$}
& \multirow{2}{*}{\makecell{\textbf{Time (s)} $\downarrow$}} \\
\cmidrule(lr){7-8}
& & & & & & \textbf{GPT-5} & \textbf{Human} & \\
\midrule
Affordance + Optimization & 3.22 & 0.73 & 8.87 & 49.15 & 0.83 & 5.57 & 5.10 & 2.84 \\
\textbf{\method{}} & \textbf{1.31} & \textbf{0.41} & \textbf{1.20} & \textbf{71.65} & \textbf{1.47} & \textbf{7.37} & \textbf{7.68} & \textbf{0.93} \\
\bottomrule
\end{tabular}
}
\caption{Ablation comparing our direct pose generation against an affordance-then-optimization baseline that shares the same predicted affordance. Our method achieves better grasp quality across all metrics while requiring far less inference time.}
\label{tab:ablation_pose_gen}
\end{table}

\clearpage
\subsection{Affordance Prediction Results}
\label{sec:supp_affordance}

\begin{figure}[H]
\centering
\includegraphics[width=\textwidth]{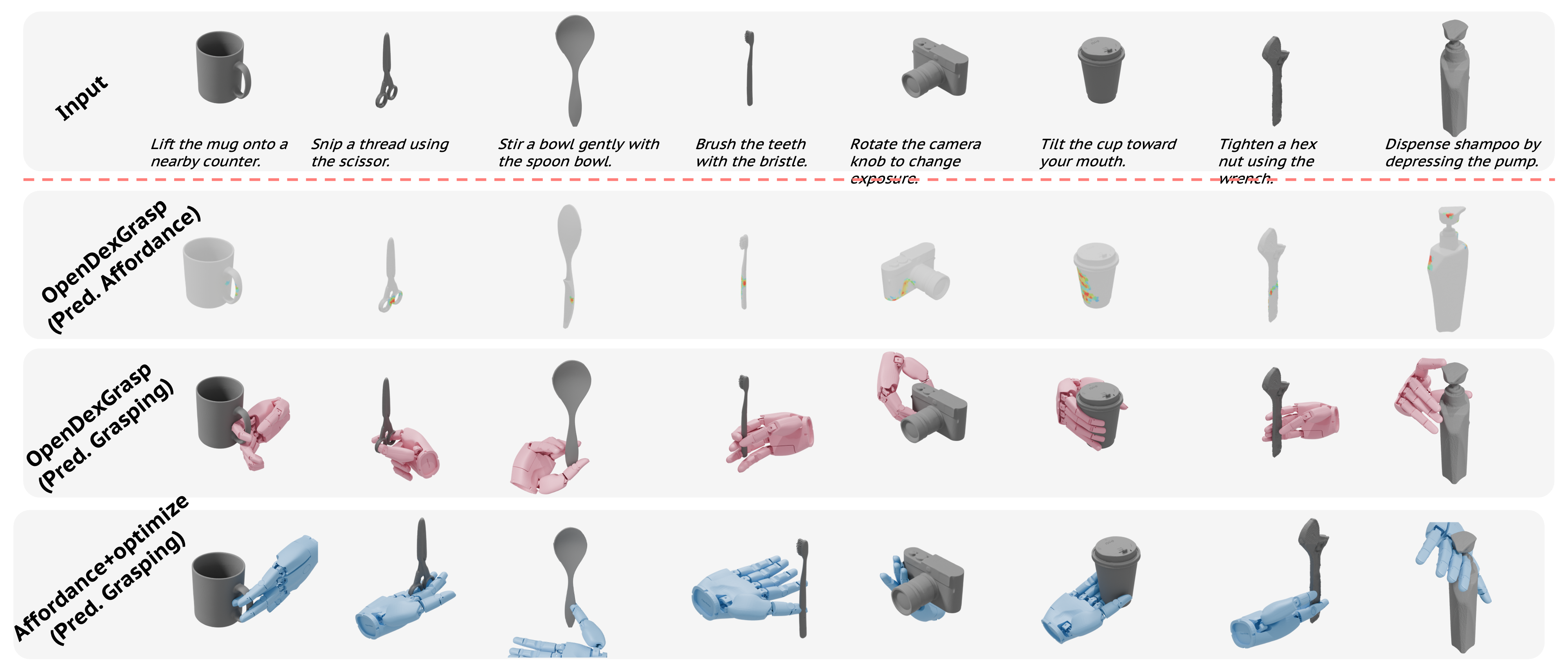}
\caption{\textbf{Instruction-conditioned grasp prediction comparison.} For each input object and a distinct language instruction, \method{} directly generates a grasp pose and affordance map over the object surface, while the affordance-then-optimization baseline optimizes a grasp from the affordance map. Our grasps remain physically plausible and aligned with the instruction-conditioned affordance, whereas the optimization baseline often produces penetrating or functionally misaligned configurations.}
\label{fig:supp_affordance}
\end{figure}

In \method{}, point-level affordance supervision serves as an auxiliary training objective that grounds the shared latent representation in functional object regions. At inference, the grasp generator is conditioned on the shared semantic and geometric representation and does not take the predicted affordance map as input. Figure~\ref{fig:supp_affordance} visualizes the auxiliary affordance predictions alongside the generated grasps and compares them with the affordance-then-optimization baseline from Section~\ref{sec:supp_ablation}. For each object--instruction pair, we show the point-level affordance map predicted by \method{}, the grasp pose directly generated from the shared representation, and the grasp produced by the optimization-based baseline. Only the baseline uses the corresponding predicted affordance map to optimize the hand configuration at test time.

Across the examples in Fig.~\ref{fig:supp_affordance}, the predicted affordance maps highlight functional regions associated with the respective instructions. The directly generated grasps exhibit finger placement and wrist orientation consistent with these regions and the intended tasks. This agreement reflects shared conditioning and auxiliary functional supervision, rather than an affordance-to-grasp inference cascade. In contrast, the optimization-based baseline exhibits physically implausible or functionally misaligned configurations in the displayed examples, including object penetration, awkward wrist poses, and contacts that drift away from the intended functional part. These qualitative results complement the ablation in Section~\ref{sec:supp_ablation}, illustrating the benefit of directly generating grasps from the shared representation over optimizing hand configurations against predicted affordance maps at test time.

\clearpage
\subsection{Real Robot Manipulation Visualization}
\label{sec:supp_robot_visualization}

\begin{figure}[H]
\centering
\includegraphics[width=\textwidth,height=0.86\textheight,keepaspectratio]{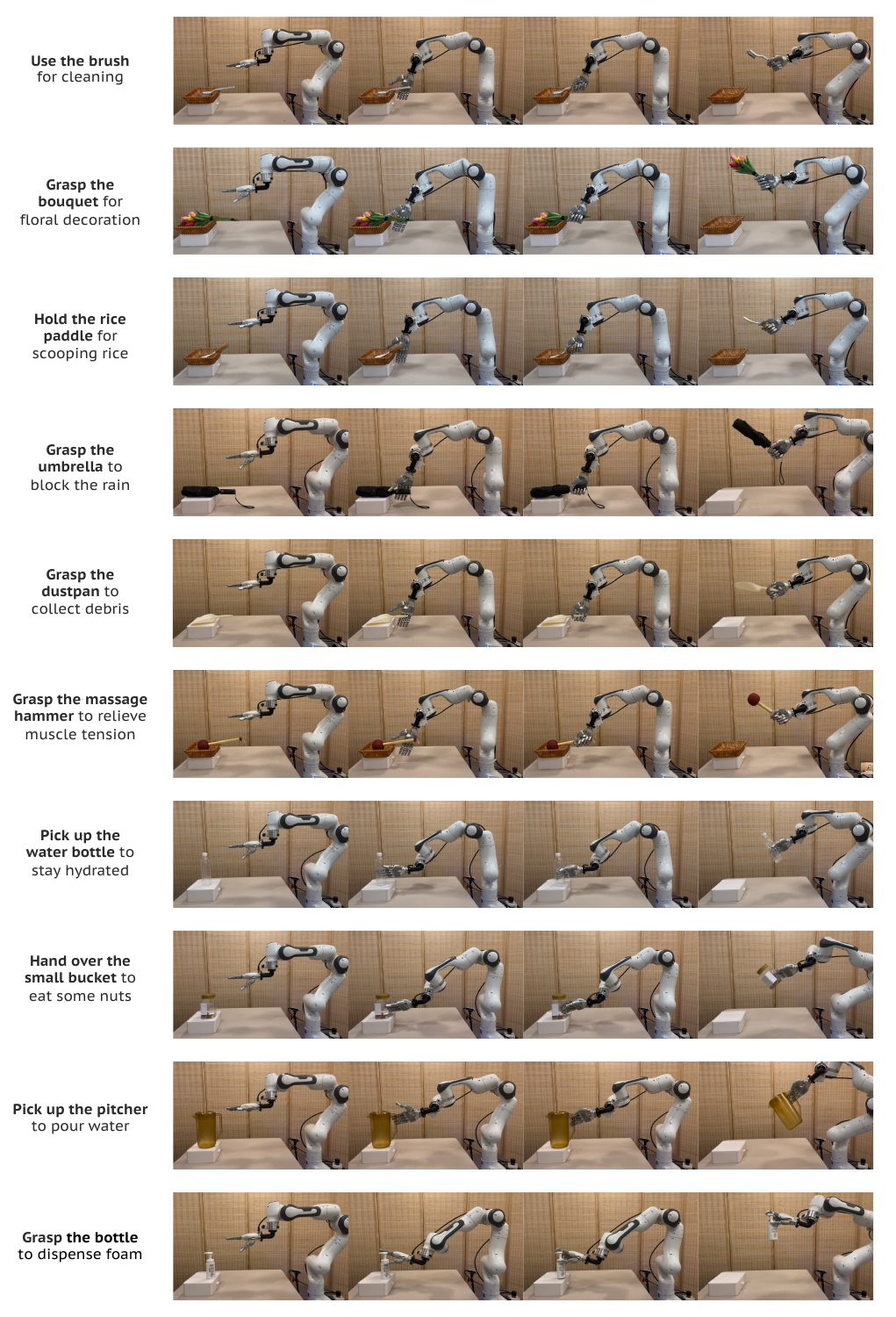}
\caption{Keyframes from ten real-world robot experiments. Each sequence shows task-conditioned dexterous grasp execution, including reaching, grasping, and lifting.}
\label{fig:real_world_exp_keyframes}
\end{figure}

Additional keyframes from ten real-world robot experiments are shown in Fig.~\ref{fig:real_world_exp_keyframes}.

%% file: Texs/15_Failure_Case_Analysis.tex
\section{Failure Case Analysis}
\label{sec:supp_failure_cases}

\begin{figure}[H]
\centering
\includegraphics[width=\textwidth]{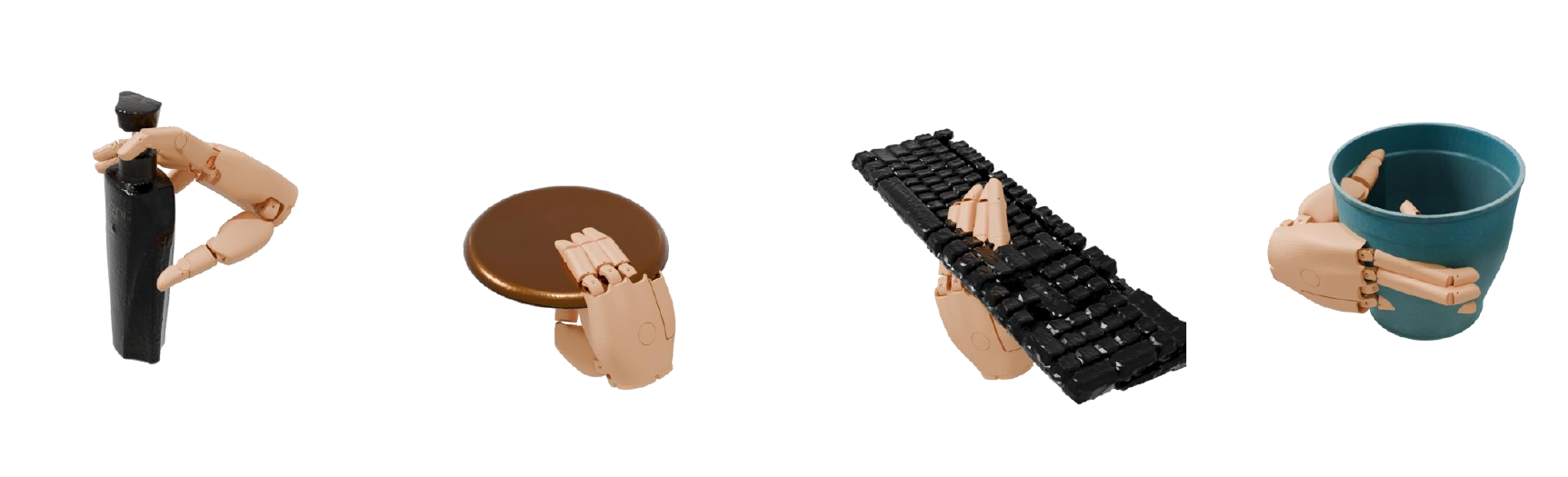}
\caption{Representative failure cases. From left to right: shampoo, frisbee, keyboard, and flower pot.}
\label{fig:failure_cases}
\end{figure}

Figure~\ref{fig:failure_cases} shows representative failure cases produced by \method{}. These examples reveal two main failure modes that are not fully captured by aggregate metrics.

\textbf{Thin-geometry and weak part-semantics failures.}
For objects such as the keyboard, frisbee, and flower pot, the dominant errors are caused by a combination of thin geometry and ambiguous part semantics. These objects contain thin plates, shells, rims, or densely structured surfaces, where a small pose error can make the hand appear visually close to a valid grasp while causing finger-object penetration. At the same time, these categories do not provide a clearly defined functional part for the instruction to ground to. As a result, the model tends to rely on a generic stable-contact prior and may place fingers on broad surfaces, rims, or key regions that are geometrically fragile for dexterous contact.

\textbf{Fine-grained functional grounding failures.}
The shampoo example illustrates a different failure mode. The predicted grasp is semantically close to the desired functional region: the hand approaches the upper part of the bottle and places the fingers near the region relevant to use. However, the fingertips are not accurately aligned with the specific functional area required by the instruction. We attribute this to the limited amount of data for fine-grained functional instructions in \humandataset{}, especially instructions that require precise contact placement on a small actuation region rather than a coarse object-level grasp. Thus, the model learns the approximate semantic neighborhood but still struggles to assign individual fingers to the exact contact locations needed for reliable functional execution.

Overall, these cases suggest that the remaining failure modes of \method{} are concentrated around high-precision contact reasoning. Thin objects require stronger geometry-aware collision handling, while localized functional interactions require more densely supervised demonstrations or explicit part-level grounding. Expanding \humandataset{} with more examples of fine-grained functional actions and incorporating higher-resolution part/contact constraints are promising directions for reducing these failures.